\documentclass{article} 
\usepackage{iclr2017_conference,times}
\usepackage{hyperref}
\usepackage{url}
\usepackage{amsmath}
\usepackage{amssymb}
\usepackage{mathtools}
\usepackage{bm}
\usepackage{times}
\usepackage{graphicx}
\usepackage{subcaption}
\usepackage{caption}
\usepackage{booktabs}
\usepackage{algorithm}
\usepackage[noend]{algpseudocode}
\usepackage{xparse}
\usepackage{color}
\usepackage{nicefrac}
\usepackage{siunitx}
\usepackage{multirow}
\usepackage{arydshln}
\usepackage{etoolbox}
\usepackage{enumitem}
\usepackage{courier}

\def\equationautorefname~#1\null{Equation~(#1)\null}

\makeatletter
\def\adl@drawiv#1#2#3{%
        \hskip.5\tabcolsep
        \xleaders#3{#2.5\@tempdimb #1{1}#2.5\@tempdimb}%
                #2\z@ plus1fil minus1fil\relax
        \hskip.5\tabcolsep}
\newcommand{\cdashlinelr}[1]{%
  \noalign{\vskip\aboverulesep
           \global\let\@dashdrawstore\adl@draw
           \global\let\adl@draw\adl@drawiv}
  \cdashline{#1}
  \noalign{\global\let\adl@draw\@dashdrawstore
           \vskip\belowrulesep}}
\makeatother

\makeatletter
\patchcmd{\hyper@makecurrent}{%
    \ifx\Hy@param\Hy@chapterstring
        \let\Hy@param\Hy@chapapp
    \fi
}{%
    \iftoggle{inappendix}{
        \@checkappendixparam{chapter}%
        \@checkappendixparam{section}%
        \@checkappendixparam{subsection}%
        \@checkappendixparam{subsubsection}%
        \@checkappendixparam{paragraph}%
        \@checkappendixparam{subparagraph}%
    }{}%
}{}{\errmessage{failed to patch}}

\newcommand*{\@checkappendixparam}[1]{%
    \def\@checkappendixparamtmp{#1}%
    \ifx\Hy@param\@checkappendixparamtmp
        \let\Hy@param\Hy@appendixstring
    \fi
}
\makeatletter

\newtoggle{inappendix}
\togglefalse{inappendix}

\apptocmd{\appendix}{\toggletrue{inappendix}}{}{\errmessage{failed to patch}}

\newcommand{\mat}[1]{\ensuremath{\bm{#1}}}
\renewcommand{\vec}[1]{\ensuremath{\boldsymbol{\mathbf{#1}}}}

\newcommand{\norm}[1]{\left\lVert#1\right\rVert}

\newcommand*{\zeros}{\ensuremath{\mathbf{0}}}

\newcommand{\x}[1]{\vec{x}_{#1}}
\renewcommand{\c}[1]{c_{#1}}

\renewcommand{\u}[1]{\vec{u}_{#1}}
\newcommand{\ulin}[1]{\vec{\bar{u}}_{#1}}
\newcommand{\s}[1]{\vec{s}_{#1}}

\ExplSyntaxOn
\DeclareExpandableDocumentCommand{\IfEmptyTF}{mmm}
{ \tl_if_empty:nTF {#1} {#2} {#3}
}
\ExplSyntaxOff
\DeclareDocumentCommand \y { o m } {%
    \IfNoValueTF{#1}{%
        \vec{y}_{#2}%
    }{%
        \vec{y}_{#2}^{(#1)}%
    }%
}
\DeclareDocumentCommand \ydot { o m } {%
    \IfNoValueTF{#1}{%
        \dot{\vec{y}}_{#2}%
    }{%
        \dot{\vec{y}}_{#2}^{(#1)}%
    }%
}
\DeclareDocumentCommand \ypred { o m } {%
    \IfNoValueTF{#1}{%
        \hat{\vec{y}}_{#2}%
    }{%
        \hat{\vec{y}}_{#2}^{(#1)}%
    }%
}
\DeclareDocumentCommand \ytrans { o m } {%
    \IfNoValueTF{#1}{%
        \tilde{\vec{y}}_{#2}%
    }{%
        \tilde{\vec{y}}_{#2}^{(#1)}%
    }%
}
\DeclareDocumentCommand \ytarg { o } {%
    \IfNoValueTF{#1}{%
        \vec{y}_{\circ}%
    }{%
        \vec{y}_{\circ}^{(#1)}%
    }%
}
\DeclareDocumentCommand \f { o m } {%
    \IfNoValueTF{#1}{%
        f%
    }{%
        f^{(#1)}%
    }%
    \IfEmptyTF{#2}{%
    }{%
        ({#2})
    }%
}
\DeclareDocumentCommand \ff { o m m } {%
    \IfNoValueTF{#1}{%
        f%
    }{%
        f^{(#1)}%
    }%
    \IfEmptyTF{#2}{%
    }{%
        _{#2}
    }%
    \IfEmptyTF{#3}{%
    }{%
        \left({#3}\right)
    }%
}
\DeclareDocumentCommand \enc { o m } {%
    \IfNoValueTF{#1}{%
        h_{\text{enc}}%
    }{%
        h_{\text{enc}}^{(#1)}%
    }%
    \IfEmptyTF{#2}{%
    }{%
        \left({#2}\right)
    }%
}
\DeclareDocumentCommand \dec { o m } {%
    \IfNoValueTF{#1}{%
        h_{\text{dec}}%
    }{%
        h_{\text{dec}}^{(#1)}%
    }%
    \IfEmptyTF{#2}{%
    }{%
        \left({#2}\right)
    }%
}
\DeclareDocumentCommand \conv { m } {%
    \mathcal{L}_{\text{conv}}%
    \IfEmptyTF{#1}{%
    }{%
        \left({#1}\right)
    }%
}
\DeclareDocumentCommand \deconv { m } {%
    \mathcal{L}_{\text{deconv}}%
    \IfEmptyTF{#1}{%
    }{%
        \left({#1}\right)
    }%
}
\DeclareDocumentCommand \maxpool { m } {%
    \mathcal{L}_{\text{maxpool}}%
    \IfEmptyTF{#1}{%
    }{%
        \left({#1}\right)
    }%
}

\DeclareDocumentCommand \upsample { m } {%
    \mathcal{L}_{\text{up}}%
    \IfEmptyTF{#1}{%
    }{%
        \left({#1}\right)
    }%
}
\DeclareDocumentCommand \d { m } {%
    d%
    \IfEmptyTF{#1}{%
    }{%
        ({#1})
    }%
}
\DeclareDocumentCommand \h { m } {%
    h%
    \IfEmptyTF{#1}{%
    }{%
        \left({#1}\right)
    }%
}
\DeclareDocumentCommand \M { o m } {%
   \IfEmptyTF{#2}{%
        \mat{M}
    }{%
        \mat{M}_{#2}
    }%
    \IfNoValueTF{#1}{%
    }{%
        ^{(#1)}%
    }%
}
\DeclareDocumentCommand \N { o m } {%
   \IfEmptyTF{#2}{%
        \mat{N}
    }{%
        \mat{N}_{#2}
    }%
    \IfNoValueTF{#1}{%
    }{%
        ^{(#1)}%
    }%
}
\DeclareDocumentCommand \W { o m } {%
   \IfEmptyTF{#2}{%
        \mat{W}
    }{%
        \mat{W}_{#2}
    }%
    \IfNoValueTF{#1}{%
    }{%
        ^{(#1)}%
    }%
}
\DeclareDocumentCommand \B { o m } {%
   \IfEmptyTF{#2}{%
        \mat{B}
    }{%
        \mat{B}_{#2}
    }%
    \IfNoValueTF{#1}{%
    }{%
        ^{(#1)}%
    }%
}
\DeclareDocumentCommand \J { o m } {%
   \IfEmptyTF{#2}{%
        \mat{J}
    }{%
        \mat{J}_{#2}
    }%
    \IfNoValueTF{#1}{%
    }{%
        ^{(#1)}%
    }%
}
\DeclareDocumentCommand \w { o m } {%
   \IfEmptyTF{#2}{%
        \vec{w}
    }{%
        \vec{w}_{#2}
    }%
    \IfNoValueTF{#1}{%
    }{%
        ^{(#1)}%
    }%
}
\DeclareDocumentCommand \qtheta { m } {%
   \IfEmptyTF{#1}{%
        \vec{\theta}
    }{%
        \vec{\theta}^{(#1)}
    }%
}
\DeclareDocumentCommand \qthetahat { m } {%
   \IfEmptyTF{#1}{%
        \hat{\vec{\theta}}
    }{%
        \hat{\vec{\theta}}^{(#1)}
    }%
}
\DeclareDocumentCommand \qphi { m } {%
    \phi
    \IfEmptyTF{#1}{%
    }{%
        \left({#1}\right)
    }%
}
\DeclareDocumentCommand \loss { o m } {%
    \IfNoValueTF{#1}{%
        \ell%
    }{%
        \ell^{(#1)}%
    }%
    \IfEmptyTF{#2}{%
    }{%
        ({#2})
    }%
}
\makeatother

\title{
Learning Visual Servoing with Deep Features and Fitted Q-Iteration
}

\author{Alex X. Lee${^\dagger}$, Sergey Levine${^\dagger}$, Pieter Abbeel${^\ddagger}{^\dagger}{^\S}$ \\
$^\dagger$ UC Berkeley, Department of Electrical Engineering and Computer Sciences\\
$^\ddagger$ OpenAI\\
$^\S$ International Computer Science Institute\\
\texttt{\{alexlee\textunderscore gk,svlevine,pabbeel\}@cs.berkeley.edu} \\
}

\iclrfinalcopy 

\setcitestyle{author,round,citesep={;},aysep={,},yysep={;}}

\begin{document}

\maketitle

\begin{abstract}

Visual servoing involves choosing actions that move a robot in response to observations from a camera, in order to reach a goal configuration in the world. Standard visual servoing approaches typically rely on manually designed features and analytical dynamics models, which limits their generalization capability and often requires extensive application-specific feature and model engineering. In this work, we study how learned visual features, learned predictive dynamics models, and reinforcement learning can be combined to learn visual servoing mechanisms. We focus on target following, with the goal of designing algorithms that can learn a visual servo using low amounts of data of the target in question, to enable quick adaptation to new targets. Our approach is based on servoing the camera in the space of learned visual features, rather than image pixels or manually-designed keypoints. We demonstrate that standard deep features, in our case taken from a model trained for object classification, can be used together with a bilinear predictive model to learn an effective visual servo that is robust to visual variation, changes in viewing angle and appearance, and occlusions. A key component of our approach is to use a sample-efficient fitted Q-iteration algorithm to learn which features are best suited for the task at hand. We show that we can learn an effective visual servo on a complex synthetic car following benchmark using just 20 training trajectory samples for reinforcement learning. We demonstrate substantial improvement over a conventional approach based on image pixels or hand-designed keypoints, and we show an improvement in sample-efficiency of more than two orders of magnitude over standard model-free deep reinforcement learning algorithms. Videos are available at \url{http://rll.berkeley.edu/visual_servoing}.

\end{abstract}

\vspace{2mm}
\section{Introduction}

Visual servoing is a classic problem in robotics that requires moving a camera or robot to match a target configuration of visual features or image intensities. Many robot control tasks that combine perception and action can be posed as visual servoing, including navigation \citep{desouza2002survey,chen2006homography}, where a robot must follow a desired path; manipulation, where the robot must servo an end-effector or a camera to a target object to grasp or manipulate it \citep{malis1999212d,corke1993servo,hashimoto1993servo,hosoda1994versatile,kragic2002survey}; and various other problems, as surveyed in \cite{hutchinson1996tutorial}. Most visual servoing methods assume access to good geometric image features \citep{chaumette2006servo,collewet2008servo,caron2013photometric} and require knowledge of their dynamics, which are typically obtained from domain knowledge about the system. Using such hand-designed features and models prevents exploitation of statistical regularities in the world, and requires manual engineering for each new system.

In this work, we study how learned visual features, learned predictive dynamics models, and reinforcement learning can be combined to learn visual servoing mechanisms. We focus on target following, with the goal of designing algorithms that can learn a visual servo using low amounts of data of the target in question, so as to be easy and quick to adapt to new targets. Successful target following requires the visual servo to tolerate moderate variation in the appearance of the target, including changes in viewpoint and lighting, as well as occlusions. Learning invariances to all such distractors typically requires a considerable amount of data. However, since a visual servo is typically specific to a particular task, it is desirable to be able to learn the servoing mechanism very quickly, using a minimum amount of data. Prior work has shown that the features learned by large convolutional neural networks on large image datasets, such as ImageNet classification \citep{deng2009imagenet}, tend to be useful for a wide range of other visual tasks \citep{donahue2014decaf}. We explore whether the usefulness of such features extends to visual servoing.

To answer this question, we propose a visual servoing method that uses pre-trained features, in our case obtained from the VGG network~\citep{simonyan2015vgg} trained for ImageNet classification. Besides the visual features, our method uses an estimate of the feature dynamics in visual space by means of a bilinear model. This allows the visual servo to predict how motion of the robot's camera will affect the perceived feature values. Unfortunately, servoing directly on the high-dimensional features of a pre-trained network is insufficient by itself to impart robustness on the servo: the visual servo must not only be robust to moderate visual variation, but it must also be able to pick out the target of interest (such as a car that the robot is tasked with following) from irrelevant distractor objects. To that end, we propose a sample-efficient fitted Q-iteration procedure that automatically chooses weights for the most relevant visual features. Crucially, the actual servoing mechanism in our approach is extremely simple, and simply seeks to minimize the Euclidean distance between the weighted feature values at the next time step and the target. The form of the servoing policy in our approach leads to an analytic and tractable linear approximator for the Q-function, which leads to a computationally efficient fitted Q-iteration algorithm. We show that we can learn an effective visual servo on a complex synthetic car following benchmark using just 20 training trajectory samples for reinforcement learning. We demonstrate substantial improvement over a conventional approach based on image pixels or hand-designed keypoints, and we show an improvement in sample-efficiency of more than two orders of magnitude over standard model-free deep reinforcement learning algorithms.

The environment for the synthetic car following benchmark is available online as the package CitySim3D\footnote{\url{https://github.com/alexlee-gk/citysim3d}},
and the code to reproduce our method and experiments is also available online\footnote{\url{https://github.com/alexlee-gk/visual_dynamics}}.
Supplementary videos of all the test executions are available on the project's website\footnote{\url{http://rll.berkeley.edu/visual_servoing}}.

\vspace{1mm}
\section{Related Work}

Visual servoing is typically (but not always) performed with calibrated cameras and carefully designed visual features. Ideal features for servoing should be stable and discriminative, and much of the work on visual servoing focuses on designing stable and convergent controllers under the assumption that such features are available \citep{espiau2002servo,mohta2014perching,wilson1996relative}. Some visual servoing methods do not require camera calibration \citep{jagersand1997experimental,yoshimi1994active}, and some recent methods operate directly on image intensities \citep{caron2013photometric}, but generally do not use learning to exploit statistical regularities in the world and improve robustness to distractors.

Learning is a relatively recent addition to the repertoire of visual servoing tools. Several methods have been proposed that apply ideas from reinforcement learning to directly acquire visual servoing controllers \citep{lampe2013acquiring,sadeghzadeh2015self}. However, such methods have not been demonstrated under extensive visual variation, and do not make use of state-of-the-art convolutional neural network visual features. Though more standard deep reinforcement learning methods \citep{lange2012autonomous,mnih2013dqn,levine2016vgps,lillicrap2016continuous} could in principle be applied to directly learn visual servoing policies, such methods tend to require large numbers of samples to learn task-specific behaviors, making them poorly suited for a flexible visual servoing algorithm that can be quickly repurposed to new tasks (e.g. to following a different object).

Instead, we propose an approach that combines learning of predictive models with pre-trained visual features. We use visual features trained for ImageNet~\citep{deng2009imagenet} classification, though any pre-trained features could in principle be applicable for our method, so long as they provide a suitable degree of invariance to visual distractors such as lighting, occlusion, and changes in viewpoint. Using pre-trained features allows us to avoid the need for large amounts of experience, but we must still learn the policy itself. To further accelerate this process, we first acquire a predictive model that allows the visual servo to determine how the visual features will change in response to an action. General video prediction is an active research area, with a number of complex but data-hungry models proposed in recent years \citep{oh2015acvp,watter2015e2c,mathieu2016multiscale,xue2016visualdynamics,lotter2017dpc,brabandere2016dfn,walker2016uncertain,vondrick2016scenedynamics}.

However, we observe that convolutional response maps can be interpreted as images and, under mild assumptions, the dynamics of image pixels during camera motion can be well approximated by means of a bilinear model \citep{censi2015jbds}.
We therefore train a relatively simple bilinear model for short-term prediction of visual feature dynamics, which we can use inside a very simple visual servo that seeks to minimize the error between the next predicted feature values and a target image.

Unfortunately, simply training predictive models on top of pre-trained features is insufficient to produce an effective visual servo, since it weights the errors of distractor objects the same amount as the object of interest. We address this challenge by using an efficient Q-iteration algorithm to train the weights on the features to maximize the servo's long-horizon reward. This method draws on ideas from regularized fitted Q-iteration~\citep{gordon1995stable,ernst2005fqi,massoud2009rfqi} and neural fitted Q-iteration~\citep{riedmiller2005nfq} to develop a sample-efficient algorithm that can directly estimate the expected return of the visual servo without the use of any additional function approximator.

\section{Problem Statement}
\label{sec:problem_statement}
Let $\y{t}$ be a featurization of the camera's observations $\x{t}$ and let $\y{*}$ be some given goal feature map.
For the purposes of this work, we define \emph{visual servoing} as the problem of choosing controls $\u{t}$ for a fixed number of discrete time steps $t$ as to minimize the error $\norm{\y{*} - \y{t}}$.

We use a relatively simple gradient-based servoing policy that uses one-step feature dynamics, ${\f{}: \{\y{t}, \u{t}\} \rightarrow \y{t+1}}$. The policy chooses the control that minimizes the distance between the goal feature map and the one-step prediction:
\begin{equation}
	\label{eq:servoing_opt}
	\pi({\x{t}, \x{*}}) = \arg\min_{\u{}} \norm{\y{*} - \f{\y{t}, \u{}}}^2.
\end{equation}
Learning this policy amounts to learning the robot dynamics and the distance metric $\norm{\cdot}$.

To learn the robot dynamics, we assume that we have access to a dataset of paired observations and controls $\x{t}, \u{t}, \x{t+1}$. This data is relatively easy to obtain as it involves collecting a stream of the robot's observations and controls. We use this dataset to learn a general visual dynamics model that can be used for any task.

To learn the distance metric, we assume that the robot interacts with the world and collects tuples of the form $\x{t}, \u{t}, \c{t}, \x{t+1}, \x{*}$. At every time step during learning, the robot observes $\x{t}$ and takes action $\u{t}$. After the transition, the robot observes $\x{t+1}$ and receives an immediate cost $\c{t}$. This cost is task-specific and it quantifies how good that transition was in order to achieve the goal. At the beginning of each trajectory, the robot is given a goal observation $\x{*}$, and it is the same throughout the trajectory. We define the goal feature map to be the featurization of the goal observation. We learn the distance metric using reinforcement learning and we model the environment as a Markov Decision Process (MDP). The state of the MDP is the tuple of the current observation and the episode's target observation, $\s{t} = (\x{t}, \x{*})$, the action $\u{t}$ is the discrete-time continuous control of the robot, and the cost function maps the states and action $(\s{t}, \u{t}, \s{t+1})$ to a scalar cost $\c{t}$.

\begin{figure}[t]
    \centering
    \begin{minipage}{.6\textwidth}
        \centering
        \includegraphics[width=\textwidth]{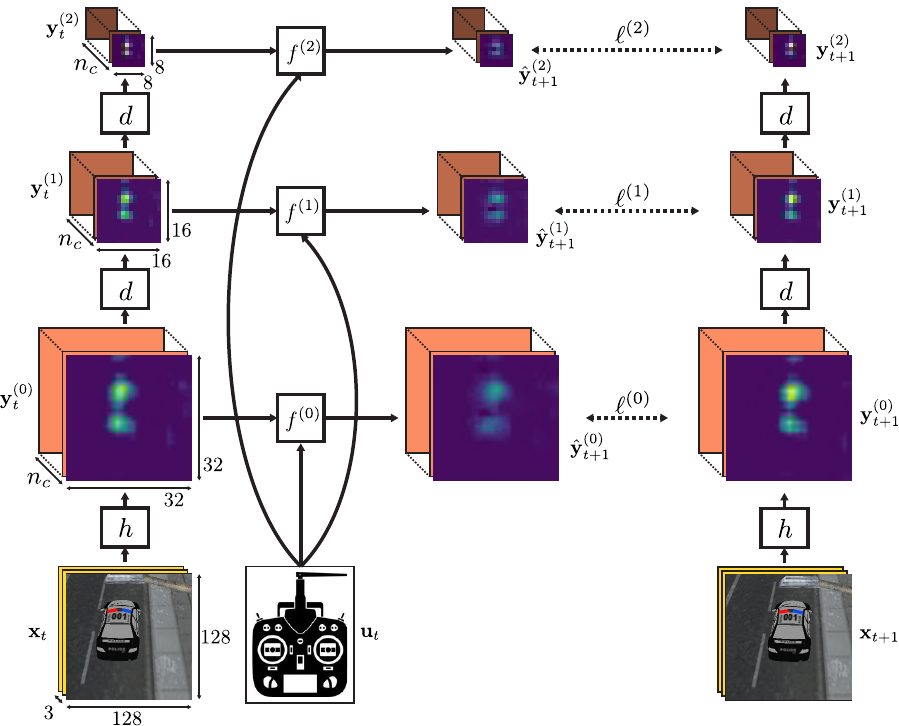}
        \caption{Multiscale bilinear model. The function $\h{}$ maps images $\x{}$ to feature maps $\y[0]{}$, the operator $\d{}$ downsamples the feature maps $\y[l-1]{}$ to $\y[l]{}$, and the bilinear function $\f[l]{}$ predicts the next feature $\ypred[l]{}$. The number of channels for each feature map is $n_c$, regardless of the scale $l$.}
        \label{fig:multiscale_bilinear_model}
    \end{minipage}
    \hfill
    \begin{minipage}{.35\textwidth}
        \centering
        \includegraphics[width=0.78\textwidth]{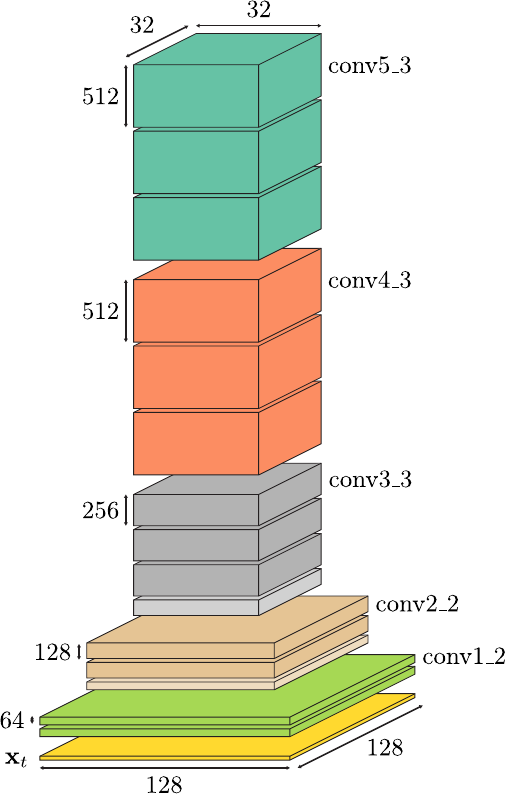}
        \caption{Dilated VGG-16 network. The intermediate feature maps drawn in a lighter shade are outputs of max-pooling layers. The features maps in the conv4 and conv5 blocks are outputs of dilated convolutions with dilation factors of 2 and 4, respectively.
        }
        \label{fig:dilated_vgg_net}
    \end{minipage}
    \vspace{-2mm}
\end{figure}

\section{Visual Features Dynamics}
\vspace{-1mm}
We learn a multiscale bilinear model to predict the visual features of the next frame given the current image from the robot's camera and the action of the robot. An overview of the model is shown in \autoref{fig:multiscale_bilinear_model}. The learned dynamics can then be used for visual servoing as described in \autoref{sec:servoing}.

\subsection{Visual Features}
\vspace{-1mm}

We consider both pixels and semantic features for the visual representation. We define the function $\h{}$ to relate the image $\x{}$ and its feature $\y{} = \h{\x{}}$. Our choice of semantic features are derived from the VGG-16 network \citep{simonyan2015vgg}, which is a convolutional neural network trained for large-scale image recognition on the ImageNet dataset \citep{deng2009imagenet}. Since spatial invariance is undesirable for servoing, we remove some of the max-pooling layers and replace the convolutions that followed them with dilated convolutions, as done by \cite{yu2016dilated}. The modified VGG network is shown in \autoref{fig:dilated_vgg_net}. We use the model weights of the original VGG-16 network, which are publicly available as a Caffe model \citep{jia2014caffe}. The features that we use are the outputs of some of the intermediate convolutional layers, that have been downsampled to a $32 \times 32$ resolution (if necessary) and standarized with respect to our training set.

We use multiple resolutions of these features for servoing. The idea is that the high-resolution representations have detailed local information about the scene, while the low-resolution representations have more global information available through the image-space gradients. The features at level $l$ of the multiscale pyramid are denoted as $\y[l]{}$. The features at each level are obtained from the features below through a downsampling operator $\d{\y[l-1]{}} = \y[l]{}$ that cuts the resolution in half.

\subsection{Bilinear Dynamics}
\label{sec:bilinear}
\vspace{-1mm}
The features $\y[l]{t}$ are used to predict the corresponding level's features $\y[l]{t+1}$ at the next time step, conditioned on the action $\u{t}$, according to a prediction function $\f[l]{\y[l]{t}, \u{t}} = \ypred[l]{t+1}$. We use a bilinear model to represent these dynamics, motivated by prior work \citep{censi2015jbds}. In order to servo at different scales, we learn a bilinear dynamics model at \emph{each} scale.
We consider two variants of the bilinear model in previous work in order to reduce the number of model parameters.

The first variant uses \emph{fully connected} dynamics as in previous work but models the dynamics of each channel independently. When semantic features are used, this model interprets the feature maps as being abstract images with spatial information within a channel and different entities or factors of variation across different channels. This could potentially allow the model to handle moving objects, occlusions, and other complex phenomena.

The fully connected bilinear model is quite large, so we propose a bilinear dynamics that enforces sparsity in the parameters. In particular, we constrain the prediction to depend only on the features that are in its local spatial neighborhood, leading to the following \emph{locally connected} bilinear model:
\setlength{\belowdisplayskip}{0pt}
\setlength{\belowdisplayshortskip}{0pt}
\begin{equation}
    \label{eq:bilinear_local}
    \ypred[l]{t+1,c} = \y[l]{t,c} + \sum_j \left( \W[l]{c,j} * \y[l]{t,c} + \B[l]{c,j} \right) \u{t,j} + \left( \W[l]{c,0} * \y[l]{t,c} + \B[l]{c,0} \right).
\end{equation}
The parameters are the 4-dimensional tensor $\W[l]{c,j}$ and the matrix $\B[l]{c,j}$ for each channel $c$, scale $l$, and control coordinate $j$.
The last two terms are biases that allow to model action-independent visual changes, such as moving objects.
The $*$ is the locally connected operator, which is like a convolution but with untied filter weights%
\footnote{
The locally connected operator, with a local neighborhood of $n_f \times n_f$ (analogous to the filter size in convolutions), is defined as:
\begin{equation*}
    \left( \W{} * \y{} \right)_{k_h, k_w} =
    \sum_{i_h = k_h - \lfloor n_f / 2\rfloor}^{k_h + \lfloor n_f / 2\rfloor}
    \sum_{i_w = k_w - \lfloor n_f / 2\rfloor}^{k_w + \lfloor n_f / 2\rfloor}
    \W{k_h, k_w, i_h - k_h, i_w - k_w} \y{i_h, i_w}.
\end{equation*}
}.

\belowdisplayskip=7pt plus 2pt minus 5pt
\belowdisplayshortskip=7pt plus 3pt minus 3pt

\subsection{Training Visual Feature Dynamics Models}
\label{sec:training}
The loss that we use for training the bilinear dynamics is the sum of the losses of the predicted features at each level, $\sum_{l=0}^{L} \loss[l]{}$, where the loss for each level $l$ is the squared $\ell$-2 norm between the predicted features and the actual features of that level, $\loss[l]{} = \lVert \y[l]{t+1} - \ypred[l]{t+1} \rVert^2$.

We optimize for the dynamics while keeping the feature representation fixed. This is a supervised learning problem, which we solve with ADAM~\citep{kingma2015adam}. The training set, consisting of triplets $\x{t}, \u{t}, \x{t+1}$, was obtained by executing a hand-coded policy that moves the robot around the target with some Gaussian noise.

\section{Learning Visual Servoing with Reinforcement Learning}
\label{sec:servoing}

We propose to use a multiscale representation of semantic features for servoing. The challenge when introducing multiple scales and multi-channel feature maps for servoing is that the features do not necessarily agree on the optimal action when the goal is unattainable or the robot is far away from the goal.
To do well, it's important to use a good weighing of each of the terms in the objective.  Since there are many weights, it would be impractically time-consuming to set them by hand, so we resort to learning.
We want the weighted one-step lookahead objective to encourage good long-term behavior, so we want this objective to correspond to the state-action value function $Q$. So we propose a method for learning the weights based on fitted Q-iteration.

\subsection{Servoing with Weighted Multiscale Features}

Instead of attempting to build an accurate predictive model for multi-step planning, we use the simple greedy servoing method in \autoref{eq:servoing_opt}, where we minimize the error between the target and predicted features for all the scales.
Typically, only a few objects in the scene are relevant, so the errors of some channels should be penalized more than others. Similarly, features at different scales might need to be weighted differently.
Thus, we use a weighting $\w[l]{c} \geq 0$ per channel $c$ and scale $l$:
\begin{equation}
    \pi(\x{t}, \x{*}) = \arg\min_{\u{}} \sum_c \sum_{l=0}^L \frac{\w[l]{c}}{|\y[l]{\cdot,c}|} \norm{\y[l]{*,c} - \ff[l]{c}{\y[l]{t,c}, \u{}}}_2^2 + \sum_j \vec{\lambda}_j \u{j}^2,
    \label{eq:servoing_opt_weighted}
\end{equation}
where $|\cdot|$ denotes the cardinality operator and the constant $\nicefrac{1}{|\y[l]{\cdot,c}|}$ normalizes the feature errors by its spatial resolution.
We also use a separate weight $\vec{\lambda}_j$ for each control coordinate $j$.
This optimization can be solved efficiently since the dynamics is linear in the controls (see \autoref{app:dynamics_linearization}).

\subsection{Q-Function Approximation for the Weighted Servoing Policy}
\vspace{-1mm}
We choose a Q-value function approximator that can represent the servoing objective such that the greedy policy with respect to the Q-values results in the policy of \autoref{eq:servoing_opt_weighted}. In particular, we use a function approximator that is linear in the weight parameters $\qtheta{}^\top = \begin{bmatrix} \w{}^\top & \vec{\lambda}^\top \end{bmatrix}$:
\begin{align*}
    Q_{\qtheta{},b}(\s{t}, \u{}) &= \phi(\s{t}, \u{})^\top \qtheta{} + b,
    &&
    \qphi{\s{t}, \u{}}^\top =
    \begin{bmatrix}
        \left\lbrack \frac{1}{|\y[l]{\cdot,c}|} \norm{\y[l]{*,c} -  \ff[l]{c}{\y[l]{t,c}, \u{}}}_2^2 \right\rbrack_{c,l}^\top &
        \left\lbrack \u{j}^2 \right\rbrack_j^\top
    \end{bmatrix}.
\end{align*}
We denote the state of the MDP as $\s{t} = (\x{t}, \x{*})$ and add a bias $b$ to the Q-function.
The servoing policy is then simply $\pi_{\qtheta{}}(\s{t}) = \arg\min_{\u{}} Q_{\qtheta{}, b}(\s{t}, \u{})$.
For reinforcement learning, we optimized for the weights $\qtheta{}$ but kept the feature representation and its dynamics fixed.

\subsection{Learning the Q-Function with Fitted Q-Iteration}
\label{sec:fqi}
\vspace{-1mm}
Reinforcement learning methods that learn a Q-function do so by minimizing the Bellman error:
\begin{equation}
    \label{eq:bellman_error}
    \norm{Q\left( \s{t}, \u{t} \right) - \left( c_t + \gamma \min_{\u{}} Q\left( \s{t+1}, \u{} \right) \right)}_2^2.
\end{equation}

In fitted Q-iteration, the agent iteratively gathers a dataset $\{\s{t}^{(i)}, \u{t}^{(i)}, \c{t}^{(i)}, \s{t+1}^{(i)}\}_i^N$ of $N$ samples according to an exploration policy, and then minimizes the Bellman error using this dataset.
We use the term \emph{sampling iteration} to refer to each iteration $j$ of this procedure.
At the beginning of each sampling iteration, the current policy with added Gaussian noise is used as the exploration policy.

It is typically hard or unstable to optimize for both Q-functions that appear in the Bellman error of \autoref{eq:bellman_error}, so it is usually optimized by iteratively optimizing the current Q-function while keeping the target Q-function constant.
However, we notice that for a given state, the action that minimizes its Q-values is the same for any non-negative scaling $\alpha$ of $\qtheta{}$ and for any bias $b$.
Thus, to speed up the optimization of the Q-function, we first set $\alpha^{(k - \frac{1}{2})}$ and $ b^{(k - \frac{1}{2})}$ by jointly solving for $\alpha$ and $b$ of \emph{both} the current and target Q-function:
\begin{equation}
    \label{eq:fqi_first_update}
    \min_{\alpha \geq 0, b} \frac{1}{N} \sum_{i=1}^N \norm{Q_{\alpha \qtheta{k-1}, b}\left( \s{t}^{(i)}, \u{t}^{(i)} \right) - \left( \c{t}^{(i)} + \gamma \min_{\u{}} Q_{\alpha \qtheta{k-1}, b}\left( \s{t+1}^{(i)}, \u{} \right) \right)}_2^2 + \nu \norm{\qtheta{}}_2^2.
\end{equation}
This is similar to how, in policy evaluation, state values can be computed by solving a linear system.
We regularize the parameters with an $\ell$-2 penalty, weighted by $\nu \geq 0$.
We use the term \emph{FQI iteration} to refer to each iteration $k$ of optimizing the Bellman error, and we use the notation $\scriptstyle (k - \frac{1}{2})$ to denote an intermediate step between iterations $\scriptstyle (k -1)$ and $\scriptstyle (k)$.
The parameters $\qtheta{}$ can then be updated with $\qtheta{k - \frac{1}{2}} = \alpha^{(k - \frac{1}{2})} \qtheta{k-1}$.
Then, we update $\qtheta{k}$ and $b^{(k)}$ by optimizing for $\qtheta{}$ and $b$ of the current Q-function while keeping the parameters of the target Q-function fixed:
\begin{equation}
    \label{eq:fqi_update}
    \min_{\qtheta{} \geq 0, b} \frac{1}{N} \sum_{i=1}^N \norm{Q_{\qtheta{}, b}\left( \s{t}^{(i)}, \u{t}^{(i)} \right) - \left( \c{t}^{(i)} + \gamma \min_{\u{}} Q_{\qtheta{k - \frac{1}{2}}, b^{(k - \frac{1}{2})}}\left( \s{t+1}^{(i)}, \u{} \right) \right)}_2^2 + \nu \norm{\qtheta{}}_2^2.
\end{equation}
A summary of the algorithm used to learn the feature weights is shown in \autoref{alg:FQI}.

\begin{algorithm}
\caption{FQI with initialization of policy-independent parameters}
\label{alg:FQI}
\begin{algorithmic}[1]
\Procedure{FQI}{$\qtheta{0}, \sigma_{\text{exploration}}^2, \nu$}
    \For{$s = 1, \dots, S$} \Comment{sampling iterations}
        \State Gather dataset $\{\s{t}^{(i)}, \u{t}^{(i)}, \c{t}^{(i)}, \s{t+1}^{(i)}\}_i^N$ using exploration policy $\mathcal{N}(\pi_{\qtheta{0}}, \sigma_{\text{exploration}}^2)$
        \For{$k = 1, \dots, K$} \Comment{FQI iterations}
            \State Fit $\alpha^{(k - \frac{1}{2})}$ and $b^{(k - \frac{1}{2})}$ using \eqref{eq:fqi_first_update}
            \State $\qtheta{k - \frac{1}{2}} \gets \alpha^{(k - \frac{1}{2})} \qtheta{k-1}$
            \State Fit $\qtheta{k}$ and $b^{(k)}$ using \eqref{eq:fqi_update}
        \EndFor
        \State $\qtheta{0} \gets \qtheta{K}$
    \EndFor
\EndProcedure
\end{algorithmic}
\end{algorithm}
\vspace{-1.3mm}

\section{Experiments}
\vspace{-1mm}
\label{sec:experiments}
We evaluate the performance of the model for visual servoing in a simulated environment. The simulated quadcopter is governed by rigid body dynamics. The robot has 4 degrees of freedom, corresponding to translation along three axis and yaw angle. This simulation is inspired by tasks in which an autonomous quadcopter flies above a city, with the goal of following some target object (e.g., a car).

\subsection{Learning Feature Dynamics and Weights with FQI}

\begin{figure}
    \centering
    \begin{minipage}{.48\textwidth}
        \centering
        \includegraphics[height=32px, trim={192px 64px 192px 160px},clip]{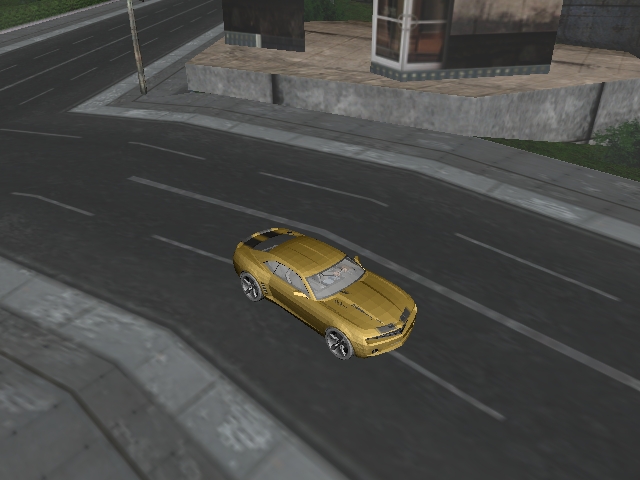}
        \includegraphics[height=32px, trim={192px 64px 192px 160px},clip]{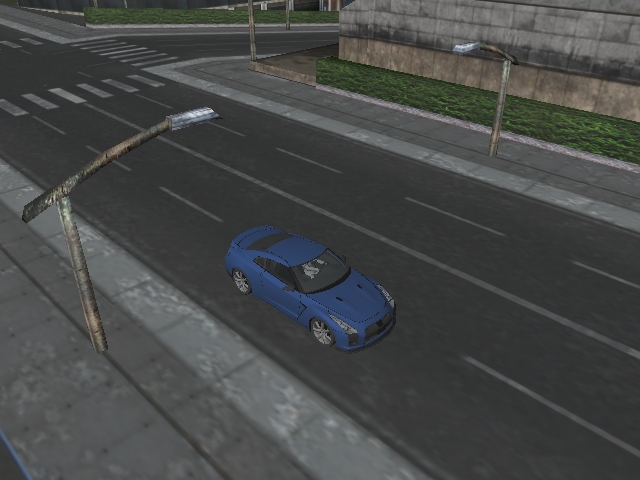}
        \includegraphics[height=32px, trim={192px 64px 192px 160px},clip]{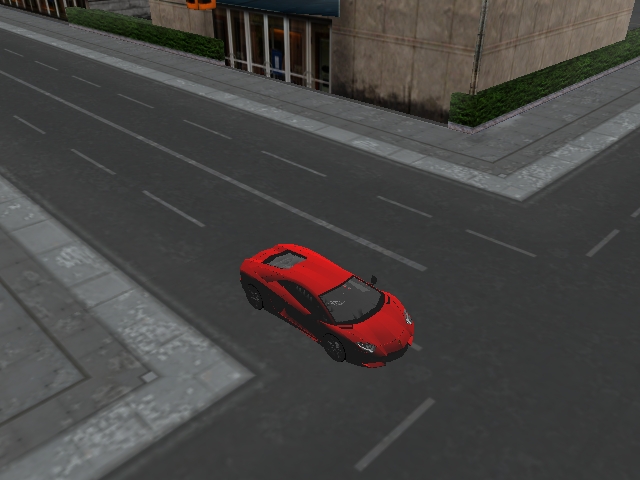}
        \includegraphics[height=32px, trim={192px 64px 192px 160px},clip]{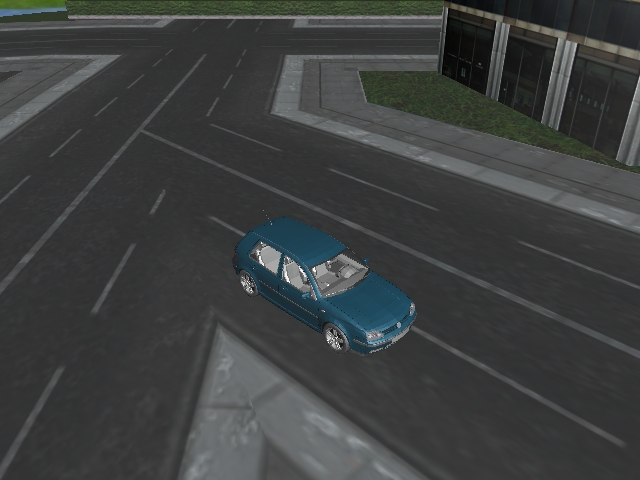}
        \includegraphics[height=32px, trim={192px 64px 192px 160px},clip]{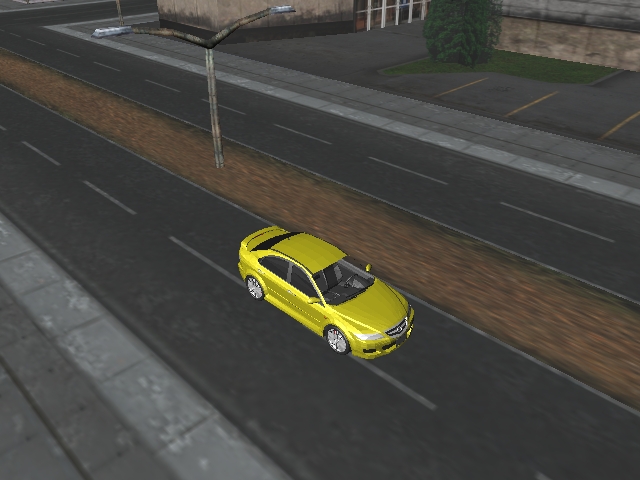}
        \caption{Cars used to learn the dynamics and the feature weights. They were also used in some of the test experiments.}
        \label{fig:train_cars}
    \end{minipage}
    \hfill
    \begin{minipage}{.48\textwidth}
        \centering
        \includegraphics[height=32px, trim={192px 64px 192px 160px},clip]{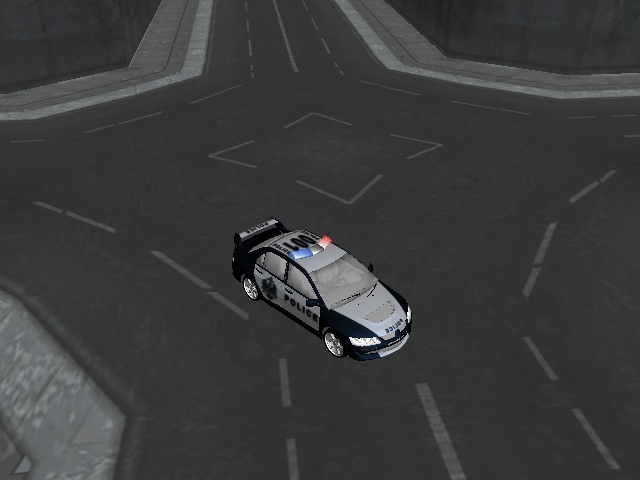}
        \includegraphics[height=32px, trim={192px 64px 192px 160px},clip]{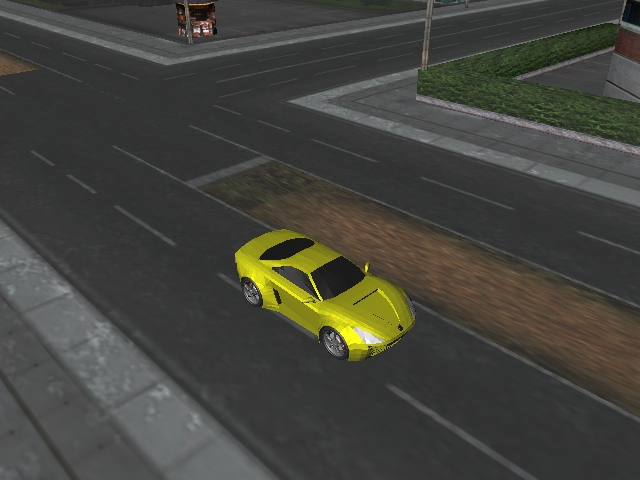}
        \includegraphics[height=32px, trim={192px 64px 192px 160px},clip]{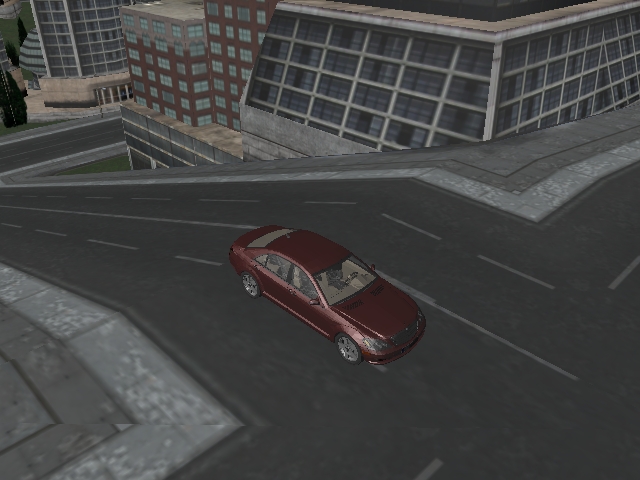}
        \includegraphics[height=32px, trim={192px 64px 192px 160px},clip]{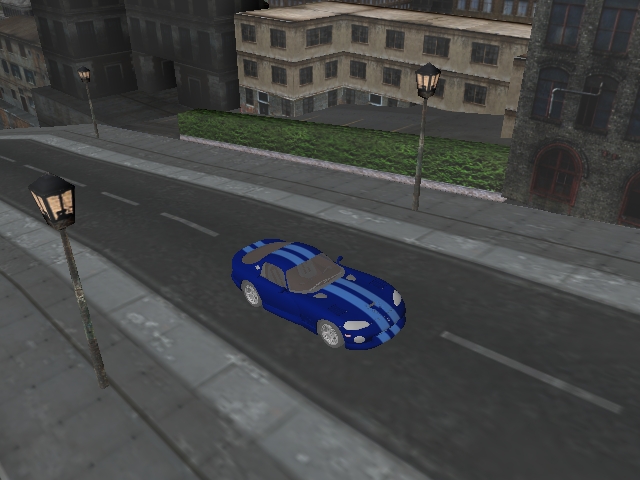}
        \includegraphics[height=32px, trim={192px 64px 192px 160px},clip]{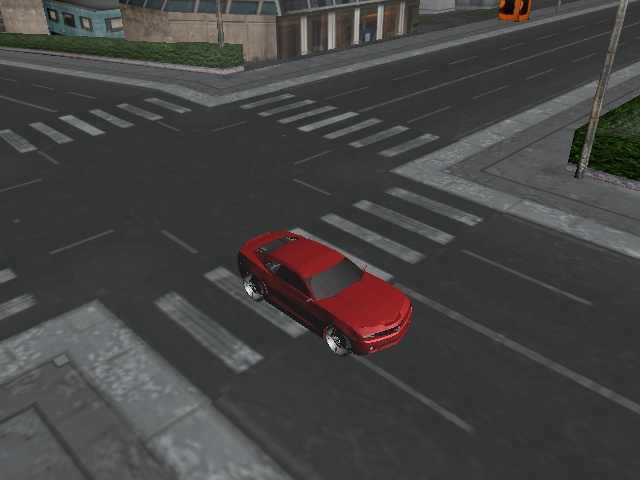}
        \caption{Novel cars used only in the test experiments. They were never seen during training or validation. \quad \quad \quad}
        \label{fig:test_cars}
    \end{minipage}
\end{figure}

The dynamics for each of the features were trained using a dataset of 10000 samples (corresponding to 100 trajectories) with ADAM~\citep{kingma2015adam}.
A single dynamics model was learned for each feature representation for all the training cars (\autoref{fig:train_cars}).
This training set was generated by executing a hand-coded policy that navigates the quadcopter around a car for 100 time steps per trajectory, while the car moves around the city.

We used the proposed FQI algorithm to learn the weightings of the features and control regularizer.
At every sampling iteration, the current policy was executed with Gaussian noise to gather data from 10 trajectories.
All the trajectories in our experiments were up to 100 time steps long.
The immediate cost received by the agent encodes the error of the target in image coordinates (details in \autoref{app:cost}).
Then, the parameters were iteratively updated by running $K = 10$ iterations of FQI.
We ran the overall algorithm for only $S = 2$ sampling iterations and chose the parameters that achieved the best performance on 10 validation trajectories. These validation trajectories were obtained by randomly choosing 10 cars from the set of training cars and randomly sampling initial states, and executing the policy with the parameters of the current iteration. All the experiments share the same set of validation trajectories.

\begin{figure}
    \centering
    \begin{subfigure}{.5\textwidth}
        \centering
        \includegraphics[width=\textwidth]{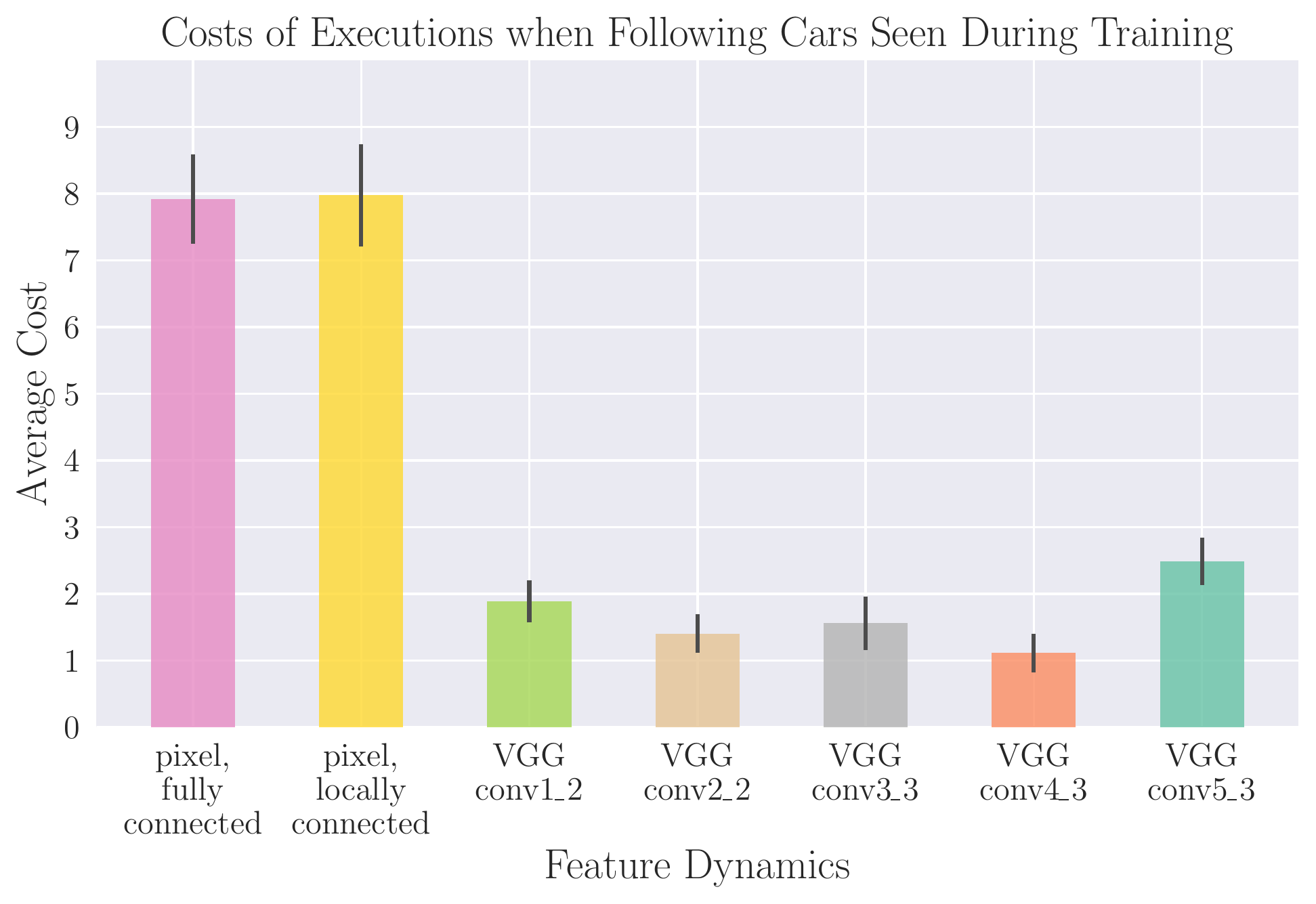}
    \end{subfigure}%
    \hfill
    \begin{subfigure}{.5\textwidth}
        \centering
        \includegraphics[width=\textwidth]{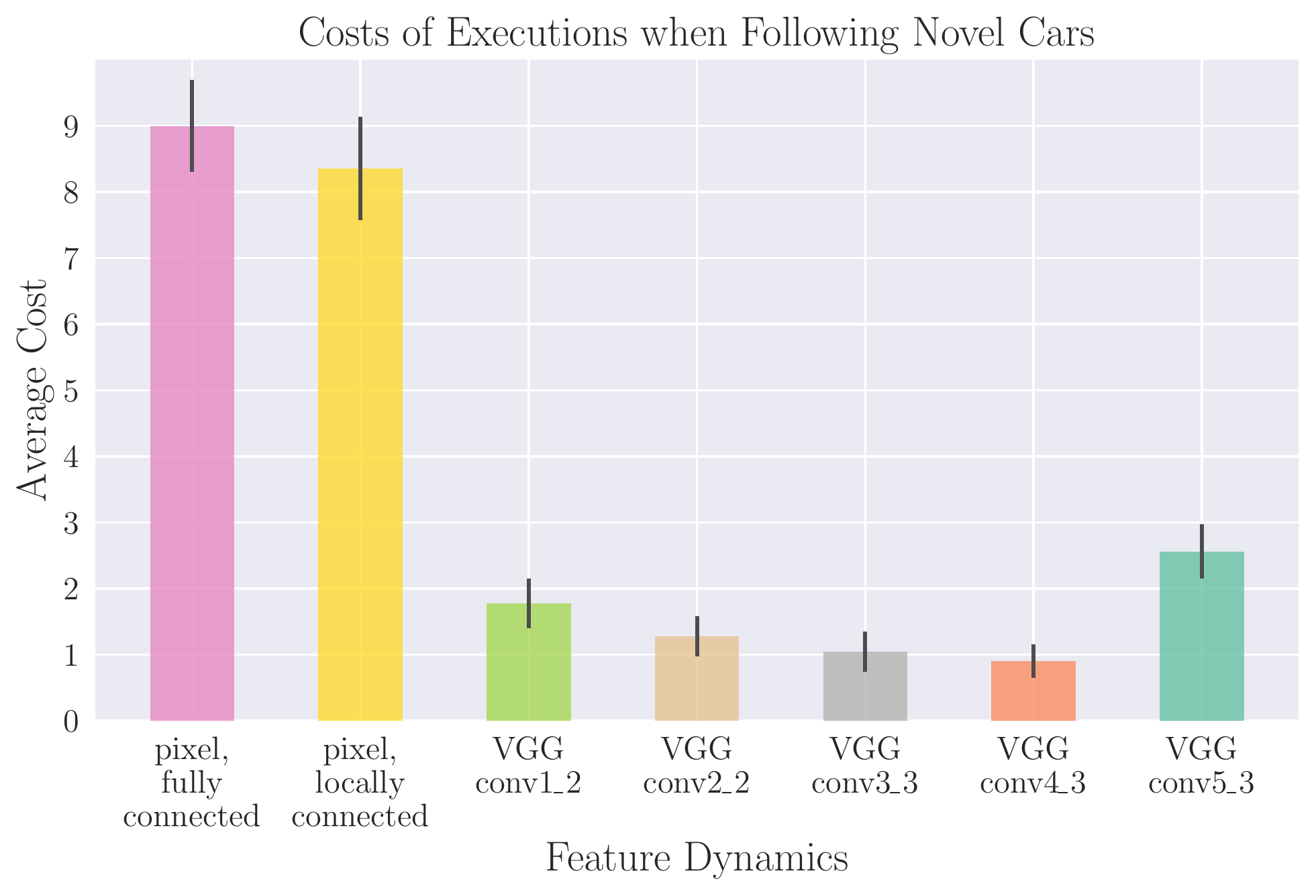}
    \end{subfigure}
    \caption{Costs of test executions using various feature dynamics models, where the feature weights are optimized with FQI.
    We test on cars that were used during learning (left plot) and on novel cars that were only used at test time (right plot).
    The reported values are the mean and standard error across 100 trajectories, of up to 100 time steps each.
    The policies based on pixel intensities use either fully connected or locally connected dynamics, whereas all the policies based on VGG features use locally connected dynamics.
    The policies based on deeper VGG features generally achieve better performance, except for the deepest feature representation, VGG conv5\textunderscore 3, which is not as suitable for approximating Q-values.
    The policies based on pixel intensities and VGG conv5\textunderscore 3 features perform worse on the novel cars. However, VGG features conv1\textunderscore 2 through conv4\textunderscore 3 achieve some degree of generalization on the novel cars.}
    \label{fig:results}
\end{figure}

\subsection{Comparison of Feature Representations for Servoing}

We compare the servoing performance for various feature dynamics models, where the weights are optimized with FQI.
We execute the learned policies on 100 test trajectories and report the average cost of the trajectory rollouts on \autoref{fig:results}. The cost of a single trajectory is the (undiscounted) sum of costs $c_t$.
We test the policies with cars that were seen during training as well as with a set of novel cars (\autoref{fig:test_cars}), to evaluate the generalization of the learned dynamics and optimized policies.

The test trajectories were obtained by randomly sampling 100 cars (with replacement) from one of the two sets of cars, and randomly sampling initial states (which are different from the ones used for validation). For consistency and reproducibility, the same sampled cars and initial states were used across all the test experiments, and the same initial states were used for both sets of cars. These test trajectories were never used during the development of the algorithm or for choosing hyperparameters.

From these results, we notice that policies based on deeper VGG features, up to VGG conv4\textunderscore 3, generally achieve better performance.
However, the deepest feature representation, VGG conv5\textunderscore 3, is not as suitable for approximating Q-values.
We hypothesize that this feature might be too spatially invariant and it might lack the necessary spatial information to differentiate among different car positions.
The policies based on pixel intensities and VGG conv5\textunderscore 3 features perform worse on the novel cars. However, VGG features conv1\textunderscore 2 through conv4\textunderscore 3 achieve some degree of generalization on the novel cars.

We show sample trajectories in \autoref{tab:trajectories_pixel_lc_vgg_conv5}. The policy based on pixel-intensities is susceptible to occlusions and distractor objects that appear in the target image or during executions. This is because distinguishing these occlusions and distractors from the cars cannot be done using just RGB features.

\begin{table}
    \centering
    \begin{tabular}{m{0.1\linewidth}m{0.75\linewidth}>{\raggedleft\arraybackslash}m{0.055\linewidth}}
        \toprule
        \begin{tabular}{@{}c@{}} Feature \\ Dynamics \end{tabular}
        & \centering Observations from Test Executions
        & \centering Cost \tabularnewline
        \midrule
        \multirow{6}{*}{\begin{tabular}{@{}l@{}l@{}} pixel, \\ locally \\ connected \end{tabular}}
        &
        \includegraphics[width=0.0825\linewidth]{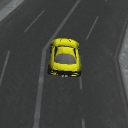}
        \includegraphics[width=0.0825\linewidth]{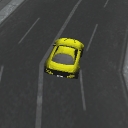}
        \includegraphics[width=0.0825\linewidth]{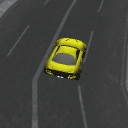}
        \includegraphics[width=0.0825\linewidth]{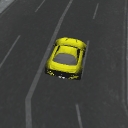}
        \includegraphics[width=0.0825\linewidth]{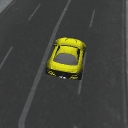}
        \includegraphics[width=0.0825\linewidth]{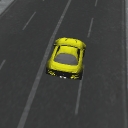}
        \includegraphics[width=0.0825\linewidth]{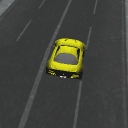}
        \includegraphics[width=0.0825\linewidth]{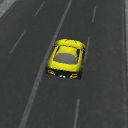}
        \includegraphics[width=0.0825\linewidth]{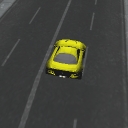}
        \includegraphics[width=0.0825\linewidth]{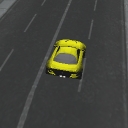}
        \includegraphics[width=0.0825\linewidth]{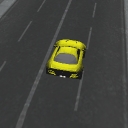}
        & 0.95
        \\
        &
        \includegraphics[width=0.0825\linewidth]{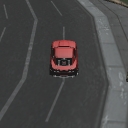}
        \includegraphics[width=0.0825\linewidth]{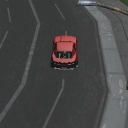}
        \includegraphics[width=0.0825\linewidth]{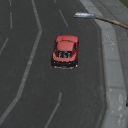}
        \includegraphics[width=0.0825\linewidth]{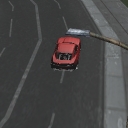}
        \includegraphics[width=0.0825\linewidth]{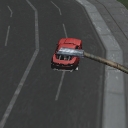}
        \includegraphics[width=0.0825\linewidth]{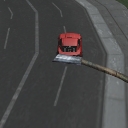}
        \includegraphics[width=0.0825\linewidth]{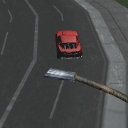}
        \includegraphics[width=0.0825\linewidth]{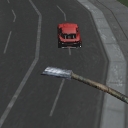}
        \includegraphics[width=0.0825\linewidth]{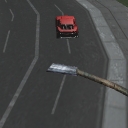}
        \includegraphics[width=0.0825\linewidth]{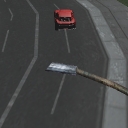}
        \includegraphics[width=0.0825\linewidth]{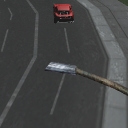}
        & 6.26
        \\
        &
        \includegraphics[width=0.0825\linewidth]{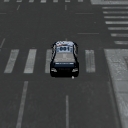}
        \includegraphics[width=0.0825\linewidth]{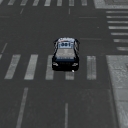}
        \includegraphics[width=0.0825\linewidth]{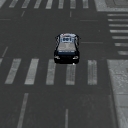}
        \includegraphics[width=0.0825\linewidth]{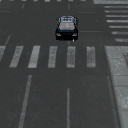}
        \includegraphics[width=0.0825\linewidth]{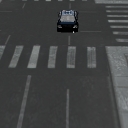}
        \includegraphics[width=0.0825\linewidth]{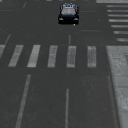}
        \includegraphics[width=0.0825\linewidth]{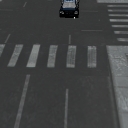}
        \includegraphics[width=0.0825\linewidth]{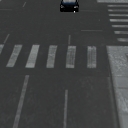}
        \includegraphics[width=0.0825\linewidth]{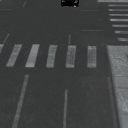}
        \includegraphics[width=0.0825\linewidth]{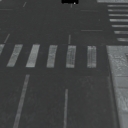}
        & 14.49
        \\ \hline \\[-1.6ex]
        \multirow{6}{*}{\begin{tabular}{@{}l@{}} VGG \\ conv4\textunderscore 3 \end{tabular}}
        &
        \includegraphics[width=0.0825\linewidth]{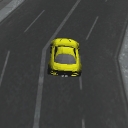}
        \includegraphics[width=0.0825\linewidth]{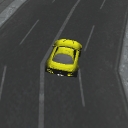}
        \includegraphics[width=0.0825\linewidth]{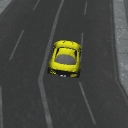}
        \includegraphics[width=0.0825\linewidth]{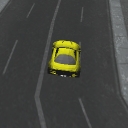}
        \includegraphics[width=0.0825\linewidth]{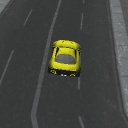}
        \includegraphics[width=0.0825\linewidth]{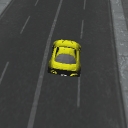}
        \includegraphics[width=0.0825\linewidth]{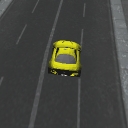}
        \includegraphics[width=0.0825\linewidth]{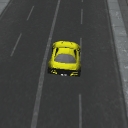}
        \includegraphics[width=0.0825\linewidth]{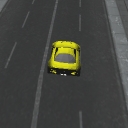}
        \includegraphics[width=0.0825\linewidth]{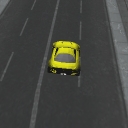}
        \includegraphics[width=0.0825\linewidth]{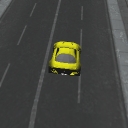}
        & 0.38
        \\
        &
        \includegraphics[width=0.0825\linewidth]{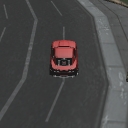}
        \includegraphics[width=0.0825\linewidth]{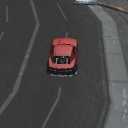}
        \includegraphics[width=0.0825\linewidth]{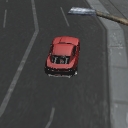}
        \includegraphics[width=0.0825\linewidth]{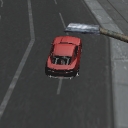}
        \includegraphics[width=0.0825\linewidth]{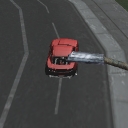}
        \includegraphics[width=0.0825\linewidth]{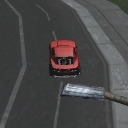}
        \includegraphics[width=0.0825\linewidth]{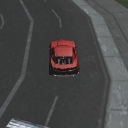}
        \includegraphics[width=0.0825\linewidth]{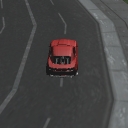}
        \includegraphics[width=0.0825\linewidth]{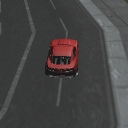}
        \includegraphics[width=0.0825\linewidth]{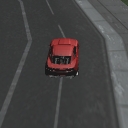}
        \includegraphics[width=0.0825\linewidth]{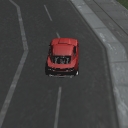}
        & 0.48
        \\
        &
        \includegraphics[width=0.0825\linewidth]{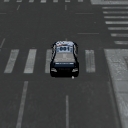}
        \includegraphics[width=0.0825\linewidth]{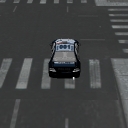}
        \includegraphics[width=0.0825\linewidth]{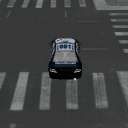}
        \includegraphics[width=0.0825\linewidth]{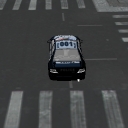}
        \includegraphics[width=0.0825\linewidth]{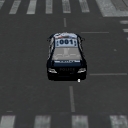}
        \includegraphics[width=0.0825\linewidth]{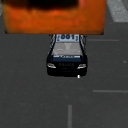}
        \includegraphics[width=0.0825\linewidth]{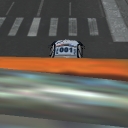}
        \includegraphics[width=0.0825\linewidth]{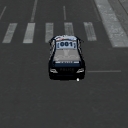}
        \includegraphics[width=0.0825\linewidth]{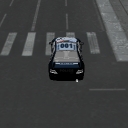}
        \includegraphics[width=0.0825\linewidth]{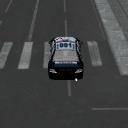}
        \includegraphics[width=0.0825\linewidth]{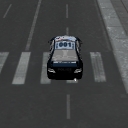}
        & 1.02
        \\
        \bottomrule
    \end{tabular}
    \caption{
    Sample observations from test executions in our experiments with the novel cars, and the costs for each trajectory, for different feature dynamics.
    We use the weights learned by our FQI algorithm. In each row, we show the observations of every 10 steps and the last one. The first observation of each trajectory is used as the target observation. The trajectories shown here were chosen to reflect different types of behaviors. The servoing policy based on pixel feature dynamics can generally follow cars that can be discriminated based on RGB pixel intensities (e.g., a yellow car with a relatively uniform background). However, it performs poorly when distractor objects appear throughout the execution (e.g., a lamp) or when they appear in the target image (e.g., the crosswalk markings on the road). On the other hand, VGG conv4\textunderscore 3 features are able to discriminate the car from distractor objects and the background, and the feature weights learned by the FQI algorithm are able to leverage this.
    Additional sample executions with other feature dynamics can be found in \autoref{tab:trajectories} in the Appendix.
    }
    \label{tab:trajectories_pixel_lc_vgg_conv5}
\end{table}

\subsection{Comparison of Weightings from Other Optimization Methods}

We compare our policy using conv4\textunderscore 3 feature dynamics, with weights optimized by FQI, against policies that use these dynamics but with either no feature weighting or weights optimized by other algorithms.

For the case of no weighting, we use a single feature weight $\w{}$ but optimize the relative weighting of the controls $\vec{\lambda}$ with the cross entropy method (CEM)~\citep{de2005cem}. For the other cases, we learn the weights with Trust Region Policy Optimization (TRPO)~\citep{schulman2015trpo}. Since the servoing policy is the minimizer of a quadratic objective (\autoref{eq:servoing_opt_weighted}), we represent the policy as a neural network that has a matrix inverse operation at the output. We train this network for 2 and 50 sampling iterations, and use a batch size of 4000 samples per iteration.
All of these methods use the same feature representation as ours, the only difference being how the weights $\w{}$ and $\vec{\lambda}$ are chosen.

\begin{figure}
    \centering
    \includegraphics[width=.6\textwidth]{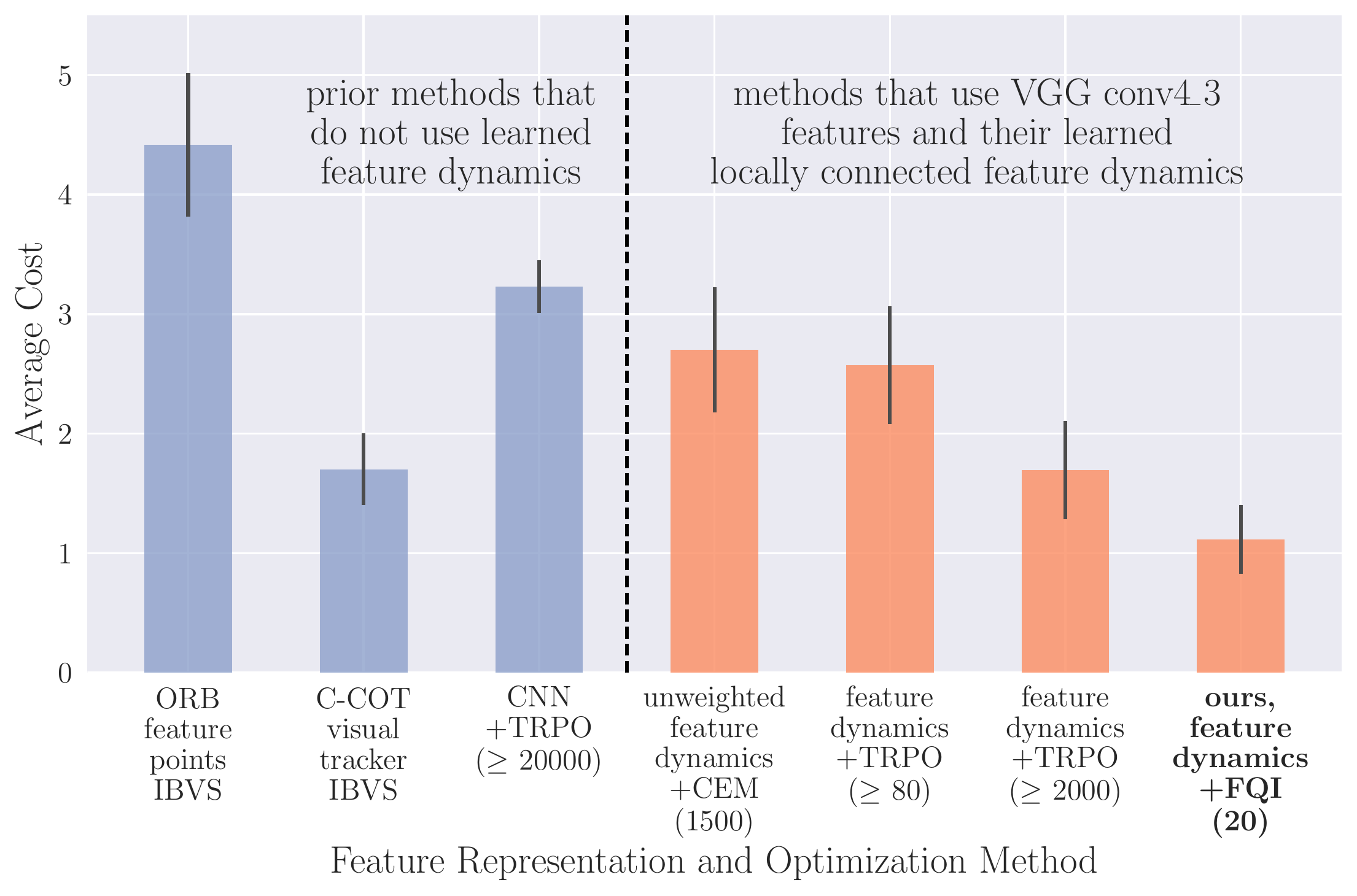}
    \caption{Comparison of costs on test executions of prior methods against our method based on VGG conv4\textunderscore 3 feature dynamics.
    These costs are from executions with the training cars; the costs are comparable when testing with the novel cars (\autoref{tab:results}).
    The first two methods use classical image-based visual servoing (IBVS) with feature points from an off-the-shelf keypoint detector and descriptor extractor (ORB features), and with feature points extracted from bounding boxes predicted by a state-of-the-art visual tracker (C-COT tracker), respectively. The third method trains a convolutional neural network (CNN) policy end-to-end with Trust Region Policy Optimization (TRPO).
    The other methods use the servoing policy based on VGG conv4\textunderscore 3 feature dynamics, either with unweighted features or weights trained with TRPO for either 2 or 50 iterations. In the case of unweighted features, we learned the weights $\vec{\lambda}$ and a single weight $\w{}$ with the cross entropy method (CEM).
    We report the number of training trajectories in parenthesis for the methods that require learning.
    For TRPO, we use a fixed number of training \emph{samples} per iteration, whereas for CEM and FQI, we use a fixed number of training \emph{trajectories} per iteration.
    We use a batch size of 4000 samples for TRPO, which means that at least 40 trajectories were used per iteration (since trajectories can terminate early, i.e. in less than 100 time steps).
    }
    \label{fig:all_results}
\end{figure}

We report the average costs of these methods on the right of \autoref{fig:all_results}. In 2 sampling iterations, the policy learned with TRPO does not improve by much, whereas our policy learned with FQI significantly outperforms the other policies. The policy learned with TRPO improves further in 50 iterations; however, the cost incurred by this policy is still about one and a half times the cost of our policy, despite using more than 100 times as many trajectories.

\subsection{Comparison to Prior Methods}

We also consider other methods that do not use the dynamics-based servoing policy that we propose. We report their average performance on the left of \autoref{fig:all_results}.

For one of the prior methods, we train a convolutional neural network (CNN) policy end-to-end with TRPO.
The policy is parametrized as a 5-layer CNN, consisting of 2 convolutional and 3 fully-connected layers, with ReLU activations except for the output layer; the convolutional layers use 16 filters ($4 \times 4$, stride 2) each and the first 2 fully-connected layers use 32 hidden units each. The policy takes in raw pixel-intensities and outputs controls.

This policy achieves a modest performance (although still worse than the policies based on conv4\textunderscore 3 feature dynamics) but it requires significantly more training samples than any of the other learning-based methods.
We also trained CNN policies that take in extracted VGG features (without any dynamics) as inputs, but they perform worse (see \autoref{tab:results_trpo} in the Appendix).
This suggests that given a policy parametrization that is expressive enough and given a large number of training samples, it is better to directly provide the raw pixel-intensity images to the policy instead of extracted VGG features.
This is because VGG features are not optimized for this task and their representation loses some information that is useful for servoing.

The other two prior methods use classical image-based visual servoing (IBVS)~\citep{chaumette2006servo} with respect to Oriented FAST and Rotated BRIEF (ORB) feature points~\citep{rublee2011orb}, or feature points extracted from a visual tracker.
For the former, the target features consist of only the ORB feature points that belong to the car, and this specifies that the car is relevant for the task.
For the tracker-based method, we use the Continuous Convolution Operator Tracker (C-COT)~\citep{danelljan2016ccot} (the current state-of-the-art visual tracker) to get bounding boxes around the car and use the four corners of the box as the feature points for servoing.
We provide the ground truth car's bounding box of the first frame as an input to the C-COT tracker.
For all of the IBVS methods, we provide the ground truth depth values of the feature points, which are used in the algorithm's interaction matrix\footnote{The term interaction matrix, or feature Jacobian, is used in the visual servo literature to denote the Jacobian of the features with respect to the control.}.

The first method performs poorly, in part because ORB features are not discriminative enough for some of the cars, and the target feature points are sometimes matched to feature points that are not on the car.
The tracker-based method achieves a relatively good performance. The gap in performance with respect to our method is in part due to the lack of car dynamics information in the IBVS model, whereas our method implicitly incorporates that in the learned feature dynamics.
It is also worth noting that the tracker-based policy runs significantly slower than our method. The open-source implementation of the C-COT tracker\footnote{https://github.com/martin-danelljan/Continuous-ConvOp} runs at about \SI[parse-numbers=false]{1}{\Hz} whereas our policy based on conv4\textunderscore 3 features runs at about \SI[parse-numbers=false]{16}{\Hz}. Most of the computation time of our method is spent computing features from the VGG network, so there is room for speedups if we use a network that is less computationally demanding.

\section{Discussion}

Manual design of visual features and dynamics models can limit the applicability of visual servoing approaches.
We described an approach that combines learned visual features with learning predictive dynamics models and reinforcement learning to learn visual servoing mechanisms.  Our experiments demonstrate that standard deep features, in our case taken from a model trained for object classification, can be used together with a bilinear predictive model to learn an effective visual servo that is robust to visual variation, changes in viewing angle and appearance, and occlusions.  For control we propose to learn Q-values, building on fitted Q-iteration, which at execution time allows for one-step lookahead calculations that optimize long term objectives.  Our method can learn an effective visual servo on a complex synthetic car following benchmark using just 20 training trajectory samples for reinforcement learning. We demonstrate substantial improvement over a conventional approach based on image pixels or hand-designed keypoints, and we show an improvement in sample-efficiency of more than two orders of magnitude over standard model-free deep reinforcement learning algorithms.

\subsubsection*{Acknowledgements}
This research was funded in part by the Army Research Office through
the MAST program, the Berkeley DeepDrive consortium, and NVIDIA. Alex Lee was
also supported by the NSF GRFP.
\bibliography{references}
\bibliographystyle{iclr2017_conference}

\appendix

\section{Linearization of the Bilinear Dynamics}
\label{app:dynamics_linearization}
The optimization of \autoref{eq:servoing_opt_weighted} can be solved efficiently by using a linearization of the dynamics,
\begin{equation}
    \ff[l]{c}{\y[l]{t,c}, \u{}} = \ff[l]{c}{\y[l]{t,c}, \ulin{}} + \J[l]{t,c} \left(\u{} - \ulin{} \right)
    = \ff[l]{c}{\y[l]{t,c}, \zeros{}} + \J[l]{t,c} \u{},
\end{equation}
where $\J[l]{t,c}$ is the Jacobian matrix with partial derivatives $\frac{\partial \ff[l]{c}{}}{\partial \u{}} (\y[l]{t,c}, \ulin{})$ and $\ulin{}$ is the linearization point. Since the bilinear dynamics are linear with respect to the controls, this linearization is exact and the Jacobian matrix does not depend on $\ulin{}$. Without loss of generality, we set $\ulin{} = \zeros{}$.

Furthermore, the bilinear dynamics allows the Jacobian matrix to be computed efficiently by simply doing a forward pass through the model. For the locally bilinear dynamics of \autoref{eq:bilinear_local}, the $j$-th column of the Jacobian matrix is given by
\begin{equation}
   \J[l]{t,c,j} = \frac{\partial \ff[l]{c}{}}{\partial \u{j}} (\y[l]{t,c}, \zeros{}) = \W[l]{c,j} * \y[l]{t,c} + \B[l]{c,j}.
\end{equation}

\section{Servoing Cost Function for Reinforcement Learning}
\label{app:cost}
The goal of reinforcement learning is to find a policy that maximizes the expected sum of rewards, or equivalently, a policy that minimizes the expected sum of costs. The cost should be one that quantifies progress towards the goal. We define the cost function in terms of the position of the target object (in the camera's local frame) after the action has been taken,
\begin{equation}
    c(\s{t}, \u{t}, \s{t+1}) =
    \begin{cases}
        \sqrt{\left(\frac{\vec{p}_{t+1}^x}{\vec{p}_{t+1}^z}\right)^2 + \left(\frac{\vec{p}_{t+1}^y}{\vec{p}_{t+1}^z}\right)^2 + \left(\frac{1}{\vec{p}_{t+1}^z} - \frac{1}{\vec{p}_{*}^z}\right)^2},
        & \text{if } ||\vec{p_{t+1}}||_2 \geq \tau \text{ and car in FOV} \\
        \left(T-t+1\right) c(\cdot, \cdot, \s{t}),
        & \text{otherwise},
    \end{cases}
\end{equation}
where $T$ is the maximum trajectory length. The episode terminates early if the camera is too close to the car (less than a distance $\tau$) or the car's origin is outside the camera's field of view (FOV). The car's position at time $t$ is $\vec{p}_t = (\vec{p}_t^x, \vec{p}_t^y, \vec{p}_t^z)$ and the car's target position is $\vec{p}_{*} = (0, 0, \vec{p}_{*}^z)$, both in the camera's local frame (z-direction is forward). Our experiments use $T=100$ and $\tau=\SI[parse-numbers=false]{4}{\meter}$.

\section{Experiment Details}
\label{app:experiment_details}

\subsection{Task Setup}
\label{app:task_setup}
The camera is attached to the vehicle slightly in front of the robot's origin and facing down at an angle of \SI[parse-numbers=false]{\nicefrac{\pi}{6}}{\radian}, similar to a commercial quadcopter drone. The robot has 4 degrees of freedom, corresponding to translation and yaw angle. Pitch and roll are held fixed.

In our simulations, the quadcopter follows a car that drives at \SI{1}{\meter\per\second} along city roads during training and testing.
The quadcopter's speed is limited to within \SI{10}{\meter\per\second} for each translational degree of freedom, and its angular speed is limited to within \SI[parse-numbers = false]{\nicefrac{\pi}{2}}{\radian\per\second}.
The simulator runs at \SI[parse-numbers = false]{10}{\Hz}.
For each trajectory, a car is chosen randomly from a set of cars, and placed randomly on one of the roads. The quadcopter is initialized right behind the car, in the desired relative position for following. The image observed at the beginning of the trajectory is used as the goal observation.

\subsection{Learning Feature Dynamics}
\label{app:learning_dynamics}
The dynamics of all the features were trained using a dataset of 10000 triplets $\x{t}, \u{t}, \x{t+1}$. The observations are $128 \times 128$ RGB images and the actions are 4-dimensional vectors of real numbers encoding the linear and angular (yaw) velocities. The actions are normalized to between $-1$ and $1$.

The training set was generated from 100 trajectories of a quadcopter following a car around the city with some randomness. Each trajectory was 100 steps long. Only 5 training cars were shown during learning. The generation process of each trajectory is as follows: First, a car is chosen at random from the set of available cars and it is randomly placed on one of the roads. Then, the quadcopter is placed at some random position relative to the car's horizontal pose, which is the car's pose that has been rotated so that the vertical axis of it and the world matches. This quadcopter position is uniformly sampled in cylindrical coordinates relative to the car's horizontal pose, with heights in the interval \SIrange{12}{18}{\meter}, and azimuthal angles in the interval \SIrange[parse-numbers = false]{-\nicefrac{\pi}{2}}{\nicefrac{\pi}{2}}{\radian} (where the origin of the azimuthal angle is the back of the car). The radii and yaw angles are initialized so that the car is in the middle of the image. At every time step, the robot takes an action that moves it towards a target pose, with some additive Gaussian noise ($\sigma = 0.2$). The target pose is sampled according to the same procedure as the initial pose, and it is sampled once at the beginning of each trajectory.

We try the fully and locally connected dynamics for pixel intensities to better understand the performance trade-offs when assuming locally connected dynamics. We do not use the latter for the semantic features since they are too high-dimensional for the dynamics model to fit in memory.
The dynamics models were trained with ADAM using 10000 iterations, a batch size of 32, a learning rate of 0.001, and momentums of 0.9 and 0.999, and a weight decay of 0.0005.

\newpage
\subsection{Learning Weighting of Feature Dynamics with Reinforcement Learning}
\label{app:reinforcement_learning}
We use CEM, TRPO and FQI to learn the feature weighting and report the performance of the learned policies in \autoref{tab:results}. We use the cost function described in \autoref{app:cost}, a discount factor of $\gamma = 0.9$, and trajectories of up to 100 steps. All the algorithms used initial weights of $\w{} = 1$ and $\vec{\lambda} = 1$, and a Gaussian exploration policy with the current policy as the mean and a fixed standard deviation $\sigma_{\text{exploration}} = 0.2$.

\setlength\tabcolsep{3pt}
\begin{table}
    \robustify\bfseries
    \centering
    \begin{subtable}{\textwidth}
        \centering
        \begin{tabular}{l
            S[separate-uncertainty=true,
              table-format=-1.2,
              table-figures-uncertainty=2]
            S[separate-uncertainty=true,
              table-format=-1.2,
              table-figures-uncertainty=2]
            S[separate-uncertainty=true,
              table-format=-2.2,
              table-figures-uncertainty=2]
            S[separate-uncertainty=true,
              table-format=-1.2,
              table-figures-uncertainty=2]
            S[separate-uncertainty=true,
              table-format=-1.2,
              table-figures-uncertainty=2,
              detect-weight]
        }
            \toprule
            & \multicolumn{5}{c}{Policy Optimization Algorithm} \\
            \cmidrule(lr){2-6}
            \multicolumn{1}{c}{\begin{tabular}{@{}c@{}} Feature \\ Dynamics \end{tabular}}
            & \multicolumn{1}{c}{\begin{tabular}{@{}c@{}c@{}c@{}} unweighted \\ feature \\ dynamics \\ + CEM (1500) \end{tabular}}
            & \multicolumn{1}{c}{\begin{tabular}{@{}c@{}c@{}c@{}} feature \\ dynamics \\ + CEM \\ (3250) \end{tabular}}
            & \multicolumn{1}{c}{\begin{tabular}{@{}c@{}c@{}c@{}} feature \\ dynamics \\ + TRPO \\ ($\geq 80$) \end{tabular}}
            & \multicolumn{1}{c}{\begin{tabular}{@{}c@{}c@{}c@{}} feature \\ dynamics \\ + TRPO \\ ($\geq 2000$) \end{tabular}}
            & \multicolumn{1}{c}{\begin{tabular}{@{}c@{}c@{}c@{}} ours, \\ feature \\ dynamics \\ + FQI (20) \end{tabular}} \\
            \midrule
            pixel, FC                  & 8.20 \pm 0.66 & 7.77 \pm 0.66 &  9.56 \pm 0.62 & 8.03 \pm 0.66 & 7.92 \pm 0.67 \\
            pixel, LC                  & 8.07 \pm 0.74 & 7.13 \pm 0.74 & 10.11 \pm 0.60 & 7.97 \pm 0.72 & 7.98 \pm 0.77 \\
            VGG conv1\textunderscore 2 & 2.22 \pm 0.38 &               &  2.06 \pm 0.35 & 1.66 \pm 0.31 & 1.89 \pm 0.32 \\
            VGG conv2\textunderscore 2 & 2.40 \pm 0.47 &               &  2.42 \pm 0.47 & 1.89 \pm 0.40 & 1.40 \pm 0.29 \\
            VGG conv3\textunderscore 3 & 2.91 \pm 0.52 &               &  2.87 \pm 0.53 & 1.59 \pm 0.42 & 1.56 \pm 0.40 \\
            VGG conv4\textunderscore 3 & 2.70 \pm 0.52 &               &  2.57 \pm 0.49 & 1.69 \pm 0.41 & \bfseries 1.11 \pm 0.29 \\
            VGG conv5\textunderscore 3 & 3.68 \pm 0.47 &               &  3.69 \pm 0.48 & 3.16 \pm 0.48 & 2.49 \pm 0.35 \\
            \bottomrule
        \end{tabular}
        \caption{Costs when using the set of cars seen during learning.}
        \label{tab:results_train_cars}
    \end{subtable}

    \bigskip
    \begin{subtable}{\textwidth}
        \centering
        \begin{tabular}{l
            S[separate-uncertainty=true,
              table-format=-1.2,
              table-figures-uncertainty=2]
            S[separate-uncertainty=true,
              table-format=-1.2,
              table-figures-uncertainty=2]
            S[separate-uncertainty=true,
              table-format=-2.2,
              table-figures-uncertainty=2]
            S[separate-uncertainty=true,
              table-format=-1.2,
              table-figures-uncertainty=2]
            S[separate-uncertainty=true,
              table-format=-1.2,
              table-figures-uncertainty=2,
              detect-weight]
        }
            \toprule
            & \multicolumn{5}{c}{Policy Optimization Algorithm} \\
            \cmidrule(lr){2-6}
            \multicolumn{1}{c}{\begin{tabular}{@{}c@{}} Feature \\ Dynamics \end{tabular}}
            & \multicolumn{1}{c}{\begin{tabular}{@{}c@{}c@{}c@{}} unweighted \\ feature \\ dynamics \\ + CEM (1500) \end{tabular}}
            & \multicolumn{1}{c}{\begin{tabular}{@{}c@{}c@{}c@{}} feature \\ dynamics \\ + CEM \\ (3250) \end{tabular}}
            & \multicolumn{1}{c}{\begin{tabular}{@{}c@{}c@{}c@{}} feature \\ dynamics \\ + TRPO \\ ($\geq 80$) \end{tabular}}
            & \multicolumn{1}{c}{\begin{tabular}{@{}c@{}c@{}c@{}} feature \\ dynamics \\ + TRPO \\ ($\geq 2000$) \end{tabular}}
            & \multicolumn{1}{c}{\begin{tabular}{@{}c@{}c@{}c@{}} ours, \\ feature \\ dynamics \\ + FQI (20) \end{tabular}} \\
            \midrule
            pixel, FC                  & 8.84 \pm 0.68 & 8.66 \pm 0.70 & 10.01 \pm 0.62 & 8.75 \pm 0.67 & 9.00 \pm 0.70 \\
            pixel, LC                  & 8.37 \pm 0.75 & 7.17 \pm 0.75 & 11.29 \pm 0.57 & 8.25 \pm 0.71 & 8.36 \pm 0.79 \\
            VGG conv1\textunderscore 2 & 2.03 \pm 0.43 &               &  1.79 \pm 0.36 & 1.42 \pm 0.33 & 1.78 \pm 0.37 \\
            VGG conv2\textunderscore 2 & 2.01 \pm 0.44 &               &  2.00 \pm 0.45 & 1.26 \pm 0.30 & 1.28 \pm 0.30 \\
            VGG conv3\textunderscore 3 & 2.03 \pm 0.47 &               &  2.08 \pm 0.47 & 1.46 \pm 0.37 & 1.04 \pm 0.31 \\
            VGG conv4\textunderscore 3 & 2.40 \pm 0.50 &               &  2.57 \pm 0.53 & 1.48 \pm 0.36 & \bfseries 0.90 \pm 0.26 \\
            VGG conv5\textunderscore 3 & 3.31 \pm 0.45 &               &  3.55 \pm 0.50 & 2.76 \pm 0.42 & 2.56 \pm 0.41 \\
            \bottomrule
        \end{tabular}
        \caption{Costs when using novel cars, none of which were seen during learning.}
        \label{tab:results_test_cars}
    \end{subtable}
    \caption{
    Costs on test executions of the dynamics-based servoing policies for different feature dynamics and weighting of the features. The reported numbers are the mean and standard error across 100 test trajectories, of up to 100 time steps each.
    We test on executions with the training cars and the novel cars; for consistency, the novel cars follow the same route as the training cars.
    We compare the performance of policies with unweighted features or weights learned by other methods.
    For the case of unweighted feature dynamics, we use the cross entropy method (CEM) to learn the relative  weights $\vec{\lambda}$ of the control and the single feature weight $\w{}$.
    For the other cases, we learn the weights with CEM, Trust Region Policy Optimization (TRPO) for either 2 or 50 iterations, and our proposed FQI algorithm. CEM searches over the full space of policy parameters $\w{}$ and $\vec{\lambda}$, but it was only ran for pixel features since it does not scale for high-dimensional problems.
    We report the number of training trajectories in parenthesis.
    For TRPO, we use a fixed number of training \emph{samples} per iteration, whereas for CEM and FQI, we use a fixed number of training \emph{trajectories} per iteration.
    We use a batch size of 4000 samples for TRPO, which means that at least 40 trajectories were used per iteration, since trajectories can terminate early, i.e. in less than 100 time steps.}
    \label{tab:results}
\end{table}

\begin{table}
    \centering
    \begin{tabular}{m{0.1\linewidth}m{0.75\linewidth}>{\raggedleft\arraybackslash}m{0.055\linewidth}}
        \toprule
        \begin{tabular}{@{}c@{}} Feature \\ Dynamics \end{tabular}
        & \centering Observations from Test Executions
        & \centering Cost \tabularnewline
        \midrule
        \multirow{3}{*}{\begin{tabular}{@{}l@{}l@{}} pixel, \\ fully \\ connected \end{tabular}}
        &
        \includegraphics[width=0.0825\linewidth]{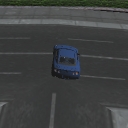}
        \includegraphics[width=0.0825\linewidth]{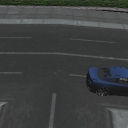}
        & 24.74
        \\
        &
        \includegraphics[width=0.0825\linewidth]{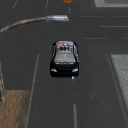}
        \includegraphics[width=0.0825\linewidth]{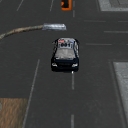}
        \includegraphics[width=0.0825\linewidth]{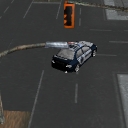}
        \includegraphics[width=0.0825\linewidth]{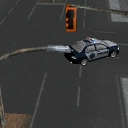}
        \includegraphics[width=0.0825\linewidth]{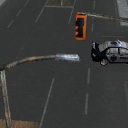}
        \includegraphics[width=0.0825\linewidth]{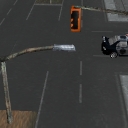}
        & 16.69
        \\ \hline \\[-1.6ex]
        \multirow{3}{*}{\begin{tabular}{@{}l@{}l@{}} pixel, \\ locally \\ connected \end{tabular}}
        &
        \includegraphics[width=0.0825\linewidth]{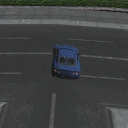}
        \includegraphics[width=0.0825\linewidth]{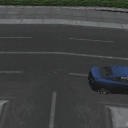}
        & 24.92
        \\
        &
        \includegraphics[width=0.0825\linewidth]{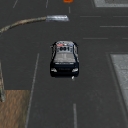}
        \includegraphics[width=0.0825\linewidth]{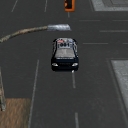}
        \includegraphics[width=0.0825\linewidth]{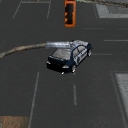}
        \includegraphics[width=0.0825\linewidth]{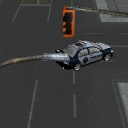}
        \includegraphics[width=0.0825\linewidth]{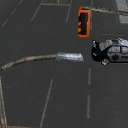}
        \includegraphics[width=0.0825\linewidth]{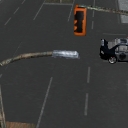}
        & 16.47
        \\ \hline \\[-1.6ex]
        \multirow{3}{*}{\begin{tabular}{@{}l@{}} VGG \\ conv1\textunderscore 2 \end{tabular}}
        &
        \includegraphics[width=0.0825\linewidth]{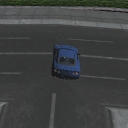}
        \includegraphics[width=0.0825\linewidth]{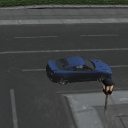}
        \includegraphics[width=0.0825\linewidth]{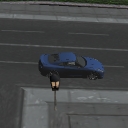}
        \includegraphics[width=0.0825\linewidth]{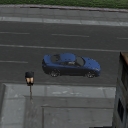}
        \includegraphics[width=0.0825\linewidth]{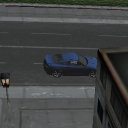}
        \includegraphics[width=0.0825\linewidth]{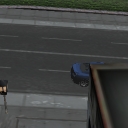}
        \includegraphics[width=0.0825\linewidth]{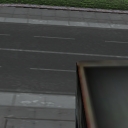}
        \includegraphics[width=0.0825\linewidth]{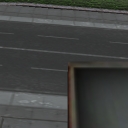}
        & 15.91
        \\
        &
        \includegraphics[width=0.0825\linewidth]{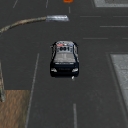}
        \includegraphics[width=0.0825\linewidth]{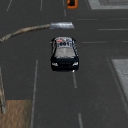}
        \includegraphics[width=0.0825\linewidth]{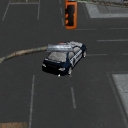}
        \includegraphics[width=0.0825\linewidth]{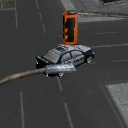}
        \includegraphics[width=0.0825\linewidth]{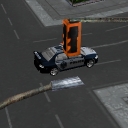}
        \includegraphics[width=0.0825\linewidth]{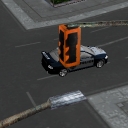}
        \includegraphics[width=0.0825\linewidth]{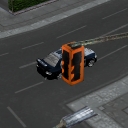}
        \includegraphics[width=0.0825\linewidth]{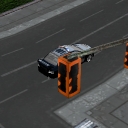}
        \includegraphics[width=0.0825\linewidth]{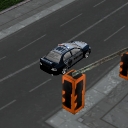}
        \includegraphics[width=0.0825\linewidth]{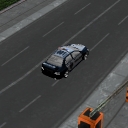}
        \includegraphics[width=0.0825\linewidth]{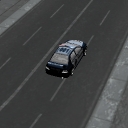}
        & 1.57
        \\ \hline \\[-1.6ex]
        \multirow{3}{*}{\begin{tabular}{@{}l@{}} VGG \\ conv2\textunderscore 2 \end{tabular}}
        &
        \includegraphics[width=0.0825\linewidth]{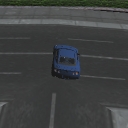}
        \includegraphics[width=0.0825\linewidth]{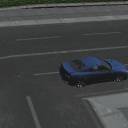}
        \includegraphics[width=0.0825\linewidth]{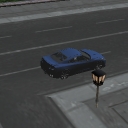}
        \includegraphics[width=0.0825\linewidth]{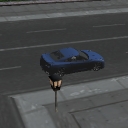}
        \includegraphics[width=0.0825\linewidth]{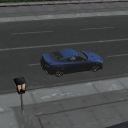}
        \includegraphics[width=0.0825\linewidth]{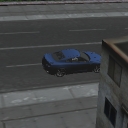}
        \includegraphics[width=0.0825\linewidth]{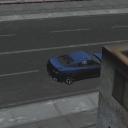}
        \includegraphics[width=0.0825\linewidth]{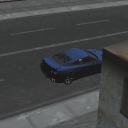}
        \includegraphics[width=0.0825\linewidth]{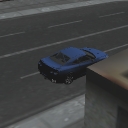}
        \includegraphics[width=0.0825\linewidth]{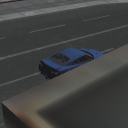}
        \includegraphics[width=0.0825\linewidth]{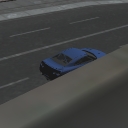}
        & 7.53
        \\
        &
        \includegraphics[width=0.0825\linewidth]{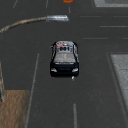}
        \includegraphics[width=0.0825\linewidth]{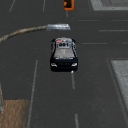}
        \includegraphics[width=0.0825\linewidth]{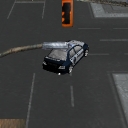}
        \includegraphics[width=0.0825\linewidth]{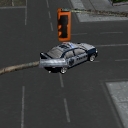}
        \includegraphics[width=0.0825\linewidth]{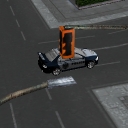}
        \includegraphics[width=0.0825\linewidth]{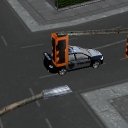}
        \includegraphics[width=0.0825\linewidth]{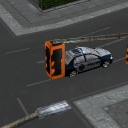}
        \includegraphics[width=0.0825\linewidth]{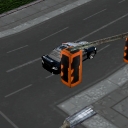}
        \includegraphics[width=0.0825\linewidth]{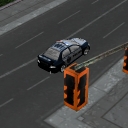}
        \includegraphics[width=0.0825\linewidth]{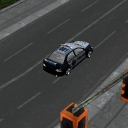}
        \includegraphics[width=0.0825\linewidth]{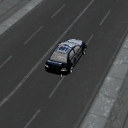}
        & 2.56
        \\ \hline \\[-1.6ex]
        \multirow{3}{*}{\begin{tabular}{@{}l@{}} VGG \\ conv3\textunderscore 3 \end{tabular}}
        &
        \includegraphics[width=0.0825\linewidth]{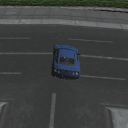}
        \includegraphics[width=0.0825\linewidth]{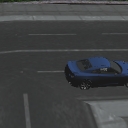}
        \includegraphics[width=0.0825\linewidth]{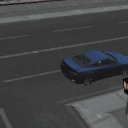}
        \includegraphics[width=0.0825\linewidth]{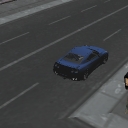}
        \includegraphics[width=0.0825\linewidth]{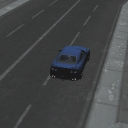}
        \includegraphics[width=0.0825\linewidth]{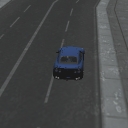}
        \includegraphics[width=0.0825\linewidth]{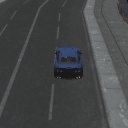}
        \includegraphics[width=0.0825\linewidth]{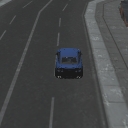}
        \includegraphics[width=0.0825\linewidth]{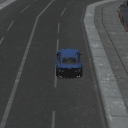}
        \includegraphics[width=0.0825\linewidth]{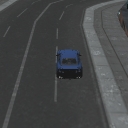}
        \includegraphics[width=0.0825\linewidth]{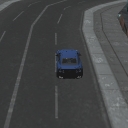}
        & 6.01
        \\
        &
        \includegraphics[width=0.0825\linewidth]{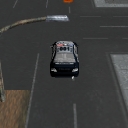}
        \includegraphics[width=0.0825\linewidth]{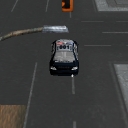}
        \includegraphics[width=0.0825\linewidth]{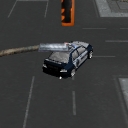}
        \includegraphics[width=0.0825\linewidth]{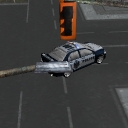}
        \includegraphics[width=0.0825\linewidth]{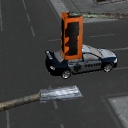}
        \includegraphics[width=0.0825\linewidth]{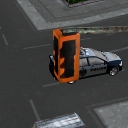}
        \includegraphics[width=0.0825\linewidth]{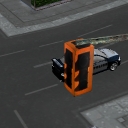}
        \includegraphics[width=0.0825\linewidth]{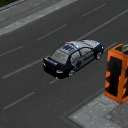}
        \includegraphics[width=0.0825\linewidth]{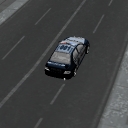}
        \includegraphics[width=0.0825\linewidth]{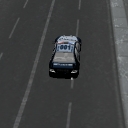}
        \includegraphics[width=0.0825\linewidth]{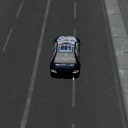}
        & 3.76
        \\ \hline \\[-1.6ex]
        \multirow{3}{*}{\begin{tabular}{@{}l@{}} VGG \\ conv4\textunderscore 3 \end{tabular}}
        &
        \includegraphics[width=0.0825\linewidth]{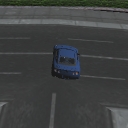}
        \includegraphics[width=0.0825\linewidth]{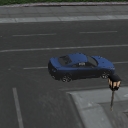}
        \includegraphics[width=0.0825\linewidth]{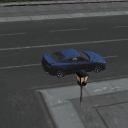}
        \includegraphics[width=0.0825\linewidth]{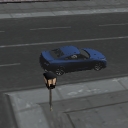}
        \includegraphics[width=0.0825\linewidth]{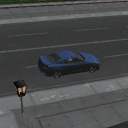}
        \includegraphics[width=0.0825\linewidth]{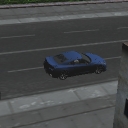}
        \includegraphics[width=0.0825\linewidth]{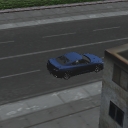}
        \includegraphics[width=0.0825\linewidth]{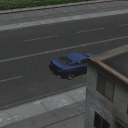}
        \includegraphics[width=0.0825\linewidth]{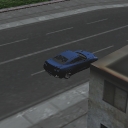}
        \includegraphics[width=0.0825\linewidth]{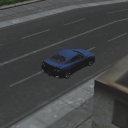}
        \includegraphics[width=0.0825\linewidth]{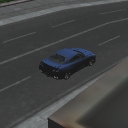}
        & 5.94
        \\
        &
        \includegraphics[width=0.0825\linewidth]{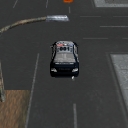}
        \includegraphics[width=0.0825\linewidth]{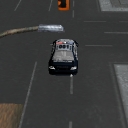}
        \includegraphics[width=0.0825\linewidth]{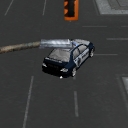}
        \includegraphics[width=0.0825\linewidth]{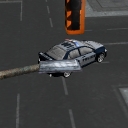}
        \includegraphics[width=0.0825\linewidth]{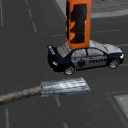}
        \includegraphics[width=0.0825\linewidth]{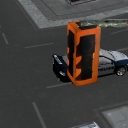}
        \includegraphics[width=0.0825\linewidth]{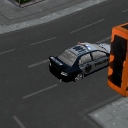}
        \includegraphics[width=0.0825\linewidth]{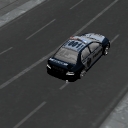}
        \includegraphics[width=0.0825\linewidth]{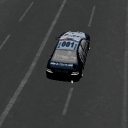}
        \includegraphics[width=0.0825\linewidth]{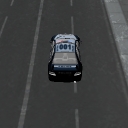}
        \includegraphics[width=0.0825\linewidth]{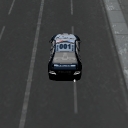}
        & 4.31
        \\ \hline \\[-1.6ex]
        \multirow{3}{*}{\begin{tabular}{@{}l@{}} VGG \\ conv5\textunderscore 3 \end{tabular}}
        &
        \includegraphics[width=0.0825\linewidth]{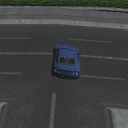}
        \includegraphics[width=0.0825\linewidth]{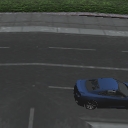}
        \includegraphics[width=0.0825\linewidth]{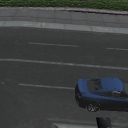}
        \includegraphics[width=0.0825\linewidth]{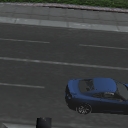}
        \includegraphics[width=0.0825\linewidth]{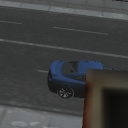}
        \includegraphics[width=0.0825\linewidth]{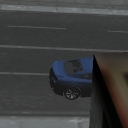}
        \includegraphics[width=0.0825\linewidth]{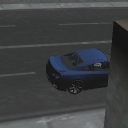}
        \includegraphics[width=0.0825\linewidth]{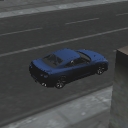}
        \includegraphics[width=0.0825\linewidth]{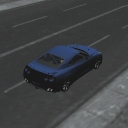}
        \includegraphics[width=0.0825\linewidth]{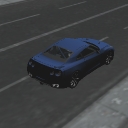}
        \includegraphics[width=0.0825\linewidth]{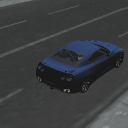}
        & 15.51
        \\
        &
        \includegraphics[width=0.0825\linewidth]{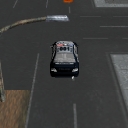}
        \includegraphics[width=0.0825\linewidth]{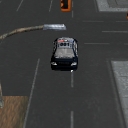}
        \includegraphics[width=0.0825\linewidth]{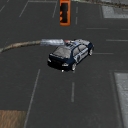}
        \includegraphics[width=0.0825\linewidth]{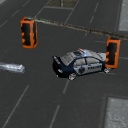}
        \includegraphics[width=0.0825\linewidth]{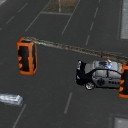}
        \includegraphics[width=0.0825\linewidth]{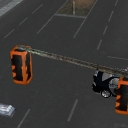}
        & 17.39
        \\
        \bottomrule
    \end{tabular}
    \caption{
    Sample observations from test executions in our experiments, and the costs for each trajectory, for different feature dynamics.
    We use the weights learned by our FQI algorithm.
    This table follows the same format as \autoref{tab:trajectories_pixel_lc_vgg_conv5}.
    Some of the trajectories were shorter than 100 steps because of the termination condition (e.g. the car is no longer in the image).
    The first observation of each trajectory is used as the target observation.
    The trajectories shown in here were chosen to reflect different types of behaviors.
    In the first trajectory, the blue car turns abruptly to the right, making the view significantly different from the target observation.
    In the second trajectory, a distractor object (i.e. the lamp) shows up in the target image and an occluder object (i.e. the traffic light) appears through the execution.
    The policies based on deeper VGG features, up to VGG conv4\textunderscore 3, are generally more robust to the appearance changes between the observations and the target observation, which are typically caused by movements of the car, distractor objects, and occlusions.}
    \label{tab:trajectories}
\end{table}

For the case of unweighted features, we use CEM to optimize for a single weight $\w{}$ and for the weights $\vec{\lambda}$. For the case of weighted features, we use CEM to optimize for the full space of parameters, but we only do that for the pixel feature dynamics since CEM does not scale for high-dimensional problems, which is the case for all the VGG features.
Each iteration of CEM performs a certain number of noisy evaluations and selects the top 20\% for the elite set. The number of noisy evaluations per iteration was 3 times the number of parameters being optimized. Each noisy evaluation used the average sum of costs of 10 trajectory rollouts as its evaluation metric.
The parameters of the last iteration were used for the final policy.
The policies with unweighted features dynamics and the policies with pixel features dynamics were trained for 10 and 25 iterations, respectively.

We use TRPO to optimize for the full space of parameters for each of the feature dynamics we consider in this work.
We use a Gaussian policy, where the mean is the servoing policy of \autoref{eq:servoing_opt_weighted} and the standard deviation is fixed to $\sigma_{\text{exploration}} = 0.2$ (i.e. we do not learn the standard deviation).
Since the parameters are constrained to be non-negative, we parametrize the TRPO policies with $\sqrt{\w{}}$ and $\sqrt{\vec{\lambda}}$.
We use a Gaussian baseline, where the mean is a 5-layer CNN, consisting of 2 convolutional and 3 fully connected layers, and a standard deviation that is initialized to $1$.
The convolutional layers use 16 filters ($4 \times 4$, stride 2) each, the first 2 fully-connected layers use 32 hidden units each, and all the layers except for the last one use ReLU activations.
The input of the baseline network are the features (either pixel intensities or VGG features) corresponding to the feature dynamics being used.
The parameters of the last iteration were used for the final policy.
The policies are trained with TRPO for 50 iterations, a batch size of 4000 samples per iteration, and a step size of 0.01.

We use our proposed FQI algorithm to optimize for the weights $\w{}, \vec{\lambda}$, and surpass the other methods in terms of performance on test executions, sample efficiency, and overall computation efficiency\footnote{Our policy based on conv4\textunderscore 3 features takes around \SI[parse-numbers=false]{650}{\s} to run $K = 10$ iterations of FQI for a given batch size of 10 training trajectories.}.
The updates of the inner iteration of our algorithm are computationally efficient; since the data is fixed for a given sampling iteration, we can precompute $\qphi{\s{t}, \u{t}}$ and certain terms of $\qphi{\s{t+1}, \cdot}$.
The parameters that achieved the best performance on 10 validation trajectories were used for the final policy.
The policies are trained with FQI for $S = 2$ sampling iterations, a batch size of 10 trajectories per sampling iteration, $K = 10$ inner iterations per sampling iteration, and a regularization coefficient of $\nu = 0.1$.
We found that regularization of the parameters was important for the algorithm to converge.
We show sample trajectories of the resulting policies in \autoref{tab:trajectories}.

\begin{figure}
    \centering
    \includegraphics[width=\textwidth]{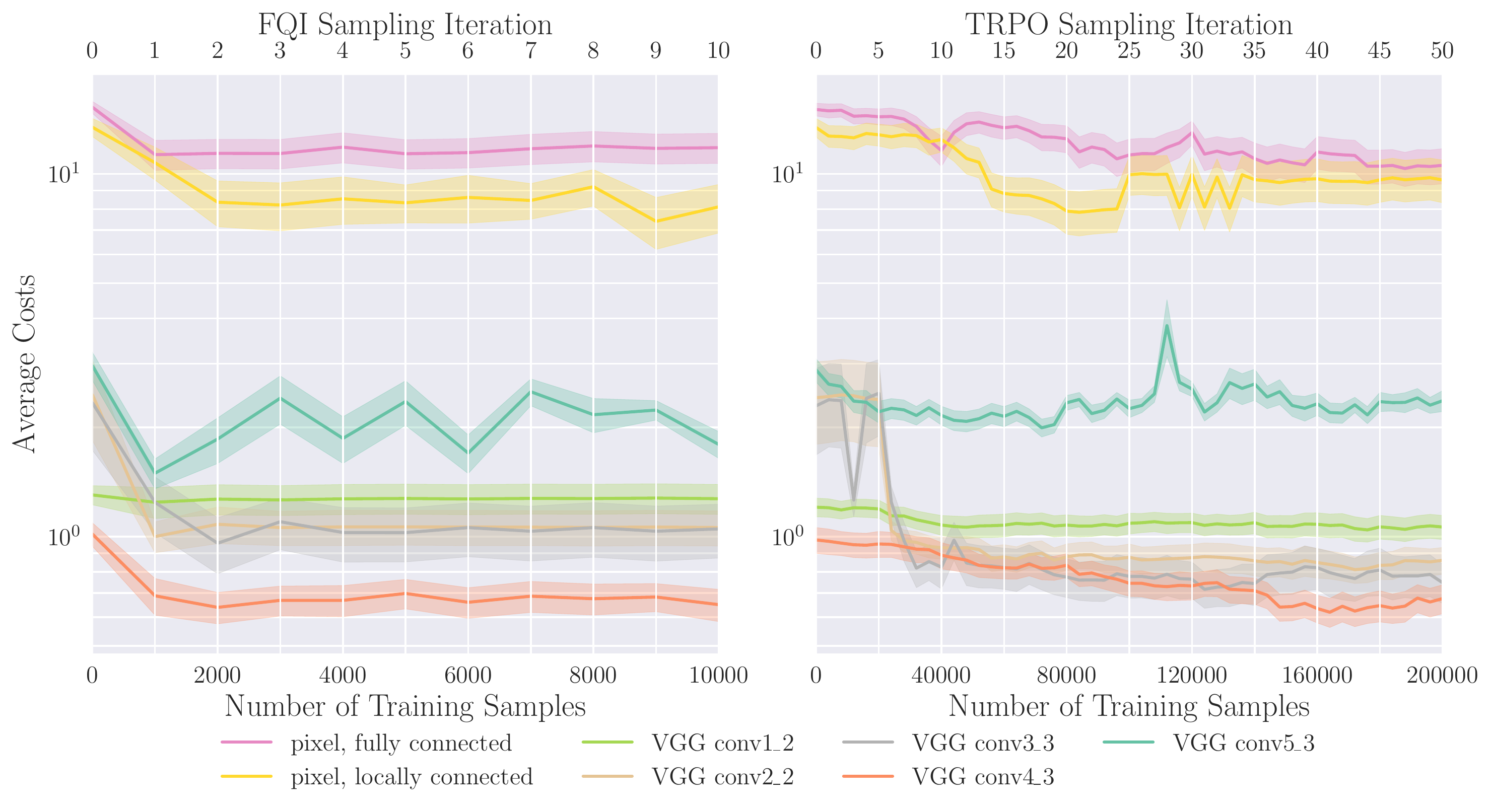}
    \caption{Costs of validation executions using various feature dynamics models, where the feature weights are optimized with FQI (left plot) or TRPO (right plot). The reported values are the mean and standard error across 10 validation trajectories, of up to 100 time steps each.}
    \label{fig:fqi_trpo_learning_val_trajs}
\end{figure}

The FQI algorithm often achieved most of its performance gain after the first iteration. We ran additional sampling iterations of FQI to see if the policies improved further. For each iteration, we evaluated the performance of the policies on 10 validation trajectories. We did the same for the policies trained with TRPO, and we compare the learning curves of both methods in \autoref{fig:fqi_trpo_learning_val_trajs}.

\subsection{Learning End-to-End Servoing Policies with TRPO}

We use TRPO to train end-to-end servoing policies for various observation modalities and report the performance of the learned policies in \autoref{tab:results_trpo}. The policies are trained with the set of training cars, and tested on both this set and on the set of novel cars.
The observation modalities that we consider are ground truth car positions (relative to the quadcopter), images of pixel intensities from the quadcopter's camera, and VGG features extracted from those images. Unlike our method and the other experiments, no feature dynamics are explicitly learned for these experiments.

We use a Gaussian policy, where the mean is either a multi-layer perceptron (MLP) or a convolutional neural net (CNN), and the standard deviation is initialized to $1$.
We also use a Gaussian baseline, which is parametrized just as the corresponding Gaussian policy (but no parameters are shared between the policy and the baseline).
For the policy that takes in car positions, the mean is parametrized as a 3-layer MLP, with tanh non-linearities except for the output layer; the first 2 fully connected layers use 32 hidden units each.
For the other policies, each of their means is parametrized as a 5-layer CNN, consisting of 2 convolutional and 3 fully-connected layers, with ReLU non-linearities except for the output layer; the convolutional layers use 16 filters ($4 \times 4$, stride 2) each and the first 2 fully-connected layers use 32 hidden units each. 

The CNN policies would often not converge for several randomly initialized parameters.
Thus, at the beginning of training, we tried multiple random seeds until we got a policy that achieved a relatively low cost on validation trajectories, and used the best initialization for training.
The MLP policy did not have this problem, so we did not have to try multiple random initializations for it.
All the policies are trained with a batch size of 4000 samples, 500 iterations, and a step size of 0.01.
The parameters of the last iteration were used for the final policy.

\begin{table}
    \centering
    \begin{subtable}{\textwidth}
        \centering
        \begin{tabular}{l
            S[separate-uncertainty=true,
              table-format=-2.2,
              table-figures-uncertainty=2]
        }
            \toprule
            \multicolumn{1}{c}{Observation Modality}
            & \\
            \midrule
            ground truth car position           &  0.59 \pm 0.24 \\
            \cdashlinelr{1-2}
            raw pixel-intensity images          &  3.23 \pm 0.22 \\
            VGG conv1\textunderscore 2 features &  7.45 \pm 0.40 \\
            VGG conv2\textunderscore 2 features & 13.38 \pm 0.53 \\
            VGG conv3\textunderscore 3 features & 10.02 \pm 0.49 \\
            \bottomrule
        \end{tabular}
        \caption{Costs when using the set of cars seen during learning.}
        \label{tab:results_trpo_train_cars}
    \end{subtable}

    \bigskip
    \begin{subtable}{\textwidth}
        \centering
        \begin{tabular}{l
            S[separate-uncertainty=true,
              table-format=-2.2,
              table-figures-uncertainty=2]
        }
            \toprule
            \multicolumn{1}{c}{Observation Modality}
            & \\
            \midrule
            ground truth car position           &  0.59 \pm 0.24 \\
            \cdashlinelr{1-2}
            raw pixel-intensity images          &  5.20 \pm 0.40 \\
            VGG conv1\textunderscore 2 features &  8.35 \pm 0.44 \\
            VGG conv2\textunderscore 2 features & 14.01 \pm 0.47 \\
            VGG conv3\textunderscore 3 features & 10.51 \pm 0.65 \\
            \bottomrule
        \end{tabular}
        \caption{Costs when using a new set of cars, none of which were seen during learning.}
        \label{tab:results_trpo_test_cars}
    \end{subtable}
    \caption{
    Costs on test executions of servoing policies that were trained end-to-end with TRPO.
    These policies take in different observation modalities: ground truth car position or image-based observations.
    This table follows the same format as \autoref{tab:results}.
    The mean of the first policy is parametrized as a 3-layer MLP, with tanh non-linearities except for the output layer; the first 2 fully connected layers use 32 hidden units each. For the other policies, each of their means is parametrized as a 5-layer CNN, consisting of 2 convolutional and 3 fully-connected layers, with ReLU non-linearities except for the output layer; the convolutional layers use 16 filters ($4 \times 4$, stride 2) each and the first 2 fully-connected layers use 32 hidden units each. All the policies are trained with TRPO, a batch size of 4000 samples, 500 iterations, and a step size of 0.01.
    The car position observations are not affected by the appearance of the cars, so the test performance for that modality is the same regardless of which set of cars are used.
    }
    \label{tab:results_trpo}
\end{table}

\subsection{Classical Image-Based Visual Servoing}
Traditional visual servoing techniques \citep{feddema1989vision,weiss1987dynamic} use the image-plane coordinates of a set of points for control. For comparison to our method, we evaluate the servoing performance of feature points derived from bounding boxes and keypoints derived from hand-engineered features, and report the costs of test executions on \autoref{tab:results_ibvs}.

We use bounding boxes from the C-COT tracker \citep{danelljan2016ccot} (the current state-of-the-art visual tracker) and ground truth bounding boxes from the simulator. The latter is defined as the box that tightly fits around the visible portions of the car.
We provide the ground truth bounding box of the first frame to the C-COT tracker to indicate that we want to track the car.
We use the four corners of the box as the feature points for servoing to take into account the position and scale of the car in image coordinates.

We provide the ground truth depth values of the feature points for the interaction matrices.
In classical image-based visual servoing, the control law involves the interaction matrix (also known as feature Jacobian), which is the Jacobian of the points in image space with respect to the camera's control (see \cite{chaumette2006servo} for details).
The analytical feature Jacobian used in IBVS assumes that the target points are static in the world frame.
This is not true for a moving car, so we consider a variant where the feature Jacobian incorporates the ground truth dynamics of the car. This amounts to adding a non-constant translation bias to the output of the dynamics function, where the translation is the displacement due to the car's movement of the 3-dimensional point in the camera's reference frame. Note that this is still not exactly equivalent to having the car being static since the roads have different slopes but the pitch and roll of the quadcopter is constrained to be fixed.

For the hand-crafted features, we consider SIFT \citep{lowe2004sift}, SURF \citep{bay2006surf} and ORB \citep{rublee2011orb} keypoints. We filter out the keypoints of the first frame that does not belong to the car and use these as the target keypoints.
However, we use all the keypoints for the subsequent observations.

The servoing policies based on bounding box features achieve low cost, and even lower ones if ground truth car dynamics is used.
However, servoing with respect to hand-crafted feature points is significantly worse than the other methods. This is, in part, because the feature extraction and matching process introduces compounding errors. Similar results were found by \cite{collewet2011photometric}, who proposed photometric visual servoing (i.e. servoing with respect to pixel intensities) and showed that it outperforms, by an order of magnitude, classical visual servoing that uses SURF features.

\begin{table}
    \centering
    \begin{tabular}{l
        S[table-space-text-pre=$\text{(0.00)}$,
          separate-uncertainty=true,
          table-format=-2.2,
          table-figures-uncertainty=2]
    }
        \toprule
        \multicolumn{1}{c}{Observation Modality (Feature Points)}
        & \\
        \midrule
        corners of bounding box from C-COT tracker              & $\text{(0.75)}$  1.70 \pm 0.30 \\
        corners of ground truth bounding box                    & $\text{(0.75)}$  0.86 \pm 0.25 \\
        corners of next frame's bounding box from C-COT tracker & $\text{(0.65)}$  1.46 \pm 0.22 \\
        corners of next frame's ground truth bounding box       & $\text{(0.65)}$  0.53 \pm 0.05 \\
        \cdashlinelr{1-2}
        SIFT feature points                                     & $\text{(0.30)}$ 14.47 \pm 0.75 \\
        SURF feature points                                     & $\text{(0.60)}$ 16.37 \pm 0.78 \\
        ORB feature points                                      & $\text{(0.30)}$  4.41 \pm 0.60 \\
        \bottomrule
    \end{tabular}
    \caption{Costs on test executions when using classical image-based visual servoing (IBVS) with respect to feature points derived from bounding boxes and keypoints derived from hand-engineered features.
    Since there is no learning involved in this method, we only test with one set of cars: the cars that were used for training in the other methods.
    This table follows the same format as \autoref{tab:results}.
    This method has one hyperparameter, which is the gain for the control law. For each feature type, we select the best hyperparameter (shown in parenthesis) by validating the policy on 10 validation trajectories for gains between 0.05 and 2, in increments of 0.05.
    The servoing policies based on bounding box features achieve low cost, and even lower ones if ground truth car dynamics is used.
    However, servoing with respect to hand-crafted feature points is significantly worse than the other methods.
    }
    \label{tab:results_ibvs}
\end{table}

\subsection{Classical Position-Based Visual Servoing}
Position-based visual servoing (PBVS) techniques use poses of a target object for control (see \cite{chaumette2006servo} for details). We evaluate the servoing performance of a few variants, and report the costs of test executions on \autoref{tab:results_pbvs}.

Similar to our IBVS experiments, we consider a variant that uses the car pose of the next time step as a way to incorporate the ground truth car dynamics into the interaction matrix.
Since the cost function is invariant to the orientation of the car, we also consider a variant where the policy only minimizes the translational part of the pose error.

These servoing policies, which use ground truth car poses, outperforms all the other policies based on images. In addition, the performance is more than two orders of magnitude better if ground truth car dynamics is used.

\begin{table}
    \centering
    \begin{tabular}{l
        S[table-space-text-pre=$\text{(0.00)}$,
          separate-uncertainty=true,
          table-format=-1.4,
          table-figures-uncertainty=4]
        S[table-space-text-pre=$\text{(0.00)}$,
          separate-uncertainty=true,
          table-format=-1.4,
          table-figures-uncertainty=4]
    }
        \toprule
        & \multicolumn{2}{c}{Policy Variant} \\
        \cmidrule(lr){2-3}
        \multicolumn{1}{c}{Observation Modality (Pose)}
        & \multicolumn{1}{c}{Use Rotation}
        & \multicolumn{1}{c}{Ignore Rotation} \\
        \midrule
        car pose              & $\text{(1.55)}$ 0.58   \pm 0.25   & $\text{(1.90)}$ 0.51   \pm 0.25   \\
        next frame's car pose & $\text{(1.00)}$ 0.0059 \pm 0.0020 & $\text{(1.00)}$ 0.0025 \pm 0.0017 \\
        \bottomrule
    \end{tabular}
    \caption{Costs on test executions when using classical position-based visual servoing (PBVS).
    Since there is no learning involved in this method, we only test with one set of cars: the cars that were used for training in the other methods.
    This table follows the same format as \autoref{tab:results}.
    This method has one hyperparameter, which is the gain for the control law. For each condition, we select the best hyperparameter (shown in parenthesis) by validating the policy on 10 validation trajectories for gains between 0.05 and 2, in increments of 0.05.
    These servoing policies, which use ground truth car poses, outperforms all the other policies based on images. In addition, the performance is more than two orders of magnitude better if ground truth car dynamics is used.
    }
    \label{tab:results_pbvs}
\end{table}

\end{document}